\documentclass[12pt]{article}
\usepackage{graphicx,psfrag,epsf}
\usepackage{enumerate}
\usepackage{natbib}
\usepackage{url} 
\usepackage{caption, subcaption}
\usepackage[margin=1in]{geometry}

\usepackage{amsmath,amsthm,amssymb,graphicx,mathtools,enumerate, textcomp, bbm, hyperref, color}
\hypersetup{colorlinks,linkcolor={blue},citecolor={blue},urlcolor={red}}  
\usepackage[title]{appendix}
\usepackage[linesnumbered,ruled,lined]{algorithm2e}
\usepackage{xcolor}
\usepackage{comment}

\newtheorem{theorem}{Theorem}[section]
\newtheorem{proposition}{Proposition}[section]

\newtheorem{lemma}[theorem]{Lemma}

\newtheorem{assumption}{}
\renewcommand*\theassumption{(A\arabic{assumption})}
\newcounter{subassumption}[assumption]
\renewcommand{\thesubassumption}{(\roman{subassumption})}
\makeatletter
\renewcommand{\p@subassumption}{\theassumption}
\makeatother
\newcommand{\subassumption}{
  \refstepcounter{subassumption}%
  \thesubassumption~\ignorespaces}

\numberwithin{equation}{section}

\newcommand{\Tra}{^{\sf T}} 

\newcommand{\R}{\mathbb{R}}

\newcommand{\E}{\mathbb{E}}

\newcommand{\Var}{\text{Var}}

\newcommand{\Ncal}{\mathcal{N}}
\newcommand{\Mcal}{\mathcal{M}}

\newcommand{\Bcal}{\mathcal{B}} 
\newcommand{\Pcal}{\mathcal{P}}
\newcommand{\Pbb}{\mathbb{P}}
\newcommand{\Lcal}{\mathcal{L}}
\newcommand{\Kcal}{\mathcal{K}}
\newcommand{\Rcal}{\mathcal{R}} 
\newcommand{\Xcal}{\mathcal{X}}
\newcommand{\Acal}{\mathcal{A}}
\newcommand{\Scal}{\mathcal{S}}
\newcommand{\Dcal}{\mathcal{D}}
\newcommand{\Fcal}{\mathcal{F}}

\newcommand{\Wcal}{\mathcal{W}}

\newcommand{\Abar}{\bar{A}}
\newcommand{\Ahat}{\hat{A}}
\newcommand{\bbar}{\bar{b}}
\newcommand{\bhat}{\hat{b}}
\newcommand{\Atilde}{\tilde{A}}
\newcommand{\btilde}{\tilde{b}}

\newcommand{\thetahat}{\hat{\theta}}
\newcommand{\thetabar}{\bar{\theta}}
\renewcommand{\qed}{\hfill$\square$}

\DeclareMathOperator*{\plim}{\mathit{p}-lim}

\DeclareMathOperator*{\argmin}{arg\,min}

\numberwithin{equation}{section}

\newcommand{\blind}{1}

\begin{document}

\def\spacingset#1{\renewcommand{\baselinestretch}%
{#1}\small\normalsize} \spacingset{1}


\if1\blind
{
  \title{\bf Online Bootstrap Inference For Policy Evaluation In Reinforcement Learning}
  \author{Pratik Ramprasad\thanks{Department of Statistics, Purdue University. Email: prampras@purdue.edu} \and
    Yuantong Li\thanks{Department of Statistics, UCLA. Email: yuantongli@g.ucla.edu} \and
    Zhuoran Yang\thanks{Department of Statistics and Data Science, Yale University. Email: zhuoran.yang@yale.edu} \and
    Zhaoran Wang\thanks{Department of Industrial Engineering and Management Sciences, Northwestern University. Email: zhaoranwang@gmail.com} \and
    Will Wei Sun\thanks{Krannert School of Management, Purdue University. Email: sun244@purdue.edu. Corresponding author. } \and
    Guang Cheng\thanks{Department of Statistics, UCLA. Email: guangcheng@stat.ucla.edu}}
  \maketitle
} \fi

\if0\blind
{
  \bigskip
  \bigskip
  \bigskip
  \begin{center}
    {\LARGE\bf Online Bootstrap Inference For Policy Evaluation in Reinforcement Learning}
\end{center}
  \bigskip
  \bigskip
  \bigskip
} \fi

\medskip
\begin{abstract}
The recent emergence of reinforcement learning (RL) has created a demand for robust statistical inference methods for the parameter estimates computed using these algorithms. Existing methods for inference in online learning are restricted to settings involving independently sampled observations, while inference methods in RL have so far been limited to the batch setting. The bootstrap is a flexible and efficient approach for statistical inference in online learning algorithms, but its efficacy in settings involving Markov noise, such as RL, has yet to be explored. In this paper, we study the use of the online bootstrap method for inference in RL policy evaluation. In particular, we focus on the temporal difference (TD) learning and Gradient TD (GTD) learning algorithms, which are themselves special instances of linear stochastic approximation under Markov noise. The method is shown to be distributionally consistent for statistical inference in policy evaluation, and numerical experiments are included to demonstrate the effectiveness of this algorithm across a range of real RL environments.
\end{abstract}

\noindent%
\vfill

\newpage
\baselineskip=24pt

\section{Introduction}
\label{sec:intro}

Reinforcement learning (RL) has achieved phenomenal success in diverse fields,  such as robotics \citep{gu2017deep}, video games \citep{mnih2015human}, autonomous driving \citep{sallab2017deep}, precision medicine \citep{parekh2019deep}, ride-sharing \citep{xu2018large}, and recommendation systems \citep{chen2019top}. A fundamental task in RL is policy evaluation, where the goal is to estimate the value function associated with a given policy from a trajectory of dependent observations. These observations may be sequentially generated either by the same policy (on-policy), or by an unknown behavior policy (off-policy). Standard algorithms used to perform on-policy and off-policy tasks include temporal difference (TD) learning \citep{sutton88} and gradient temporal difference (GTD) learning \citep{sutton2009fast}, respectively. Both are instances of linear stochastic approximation.

While RL has proven to be remarkably successful in various applications, there are still many challenges in real-world systems that prevent it from being applied at scale in practice \citep{Dulac-Arnold2021}. A primary bottleneck in real applications is that environmental interactions are prohibitively expensive. Offline RL \citep{levine2020offline} attempts to address this by training and evaluating multiple policies using pre-existing datasets collected using a single policy, thereby circumventing the need for additional environment interactions. Unfortunately, off-policy value estimates can be challenging in cases where the trained policy is substantially different from the behavior policy, i.e., the policy that was used to collect the data. In such cases, it is important to evaluate policies in an online setting prior to final deployment. \textcolor{black}{Moreover, in many real applications, we are often interested in obtaining not just the point estimate of the value function, but also a measure of the statistical uncertainty associated with the estimate. For example, online randomized experiments, e.g., A/B testing, have been widely conducted by technological/pharmaceutical companies to compare a new product with an old one. Recent studies \citep{li2021unifying, shi2021online, shi2022dynamic} have used various online updating methods to form sequential testing procedures. In these online evaluation tasks, it is important to quantify the uncertainty of the point estimate for constructing a valid hypothesis testing.  In recommender systems, a new policy is typically tested on a small fraction of user traffic in an online fashion prior to deployment. Providing a confidence interval of the predicted value estimation can help make a recommendation with more confidence because an unstable recommendation can potentially reduce users' trust in the system \citep{adomavicius2012stability, chen2019inference}. Similarly, in autonomous driving, policies must be evaluated on real test tracks prior to real-world application and it is often risky to run a policy without a statistically sound estimate of its quality.} 


Existing inference methods in RL mainly focus on off-policy evaluation using batch updates, which are often computationally inefficient in sequential data scenarios. This motivates the development of online inferential tools for policy evaluation in RL.

\color{black}
\subsubsection*{Our Contributions}
In this work, we present a \textbf{fully online} multiplier-type bootstrap algorithm that allows for statistical inference in RL policy evaluation. To the best of our knowledge, this is the first online uncertainty quantification method for sequential decision making under Markovian noise. Since our method is designed for the general framework of stochastic approximation, it applies to both the on-policy and off-policy evaluation tasks, and also to other online learning problems such as stochastic gradient descent (SGD) under Markov noise. 

\textcolor{black}{From a theoretical standpoint, our main contribution is to establish the distributional consistency of our bootstrap estimator within the Markovian noise setting. Existing inference methods for online learning \citep{fang19, chen20sgd} have mainly focused on the i.i.d. noise setting, which allows for asymptotic analysis using the martingale properties of the residual noise in the stochastic approximation update. In the present work, the Markovian noise assumption precludes the direct use of martingale central limit theory to characterize the asymptotic properties of our estimator, and necessitates a novel combination of techniques from the Markov Chain Monte-Carlo (MCMC), stochastic approximation, and RL finite-sample analysis literature. Notably, our theoretical results hold under standard assumptions from the RL literature, and do not require any kind of projection step in the stochastic approximation algorithm, despite the presence of Markovian noise.} 

Our numerical experiments demonstrate that the algorithm performs on par with the vanilla (offline) bootstrap method across a range of environments, but with substantial savings in terms of data storage and computational cost. This is especially pertinent to computationally demanding algorithms such as deep Q-learning \citep{mnih2015human}, where uncertainty quantification using the standard bootstrap method is often unfeasible due to the massive computational and storage costs involved.

To summarize, our contributions are threefold. First, we develop a fully online inference method for linear stochastic approximation under Markov noise, with applications to both on-policy (TD) and off-policy (GTD) reinforcement learning algorithms. Second, we prove that the confidence intervals constructed using our bootstrap algorithm are asymptotically valid. Third, we demonstrate the efficacy of this method through a number of numerical experiments, including on-policy evaluation with linear TD learning \citep{sutton2018reinforcement}, deep Q-learning \citep{mnih2015human}, and off-policy evaluation with GTD learning \citep{sutton08, sutton09}.

\subsubsection*{Related Work}
There has recently been a growth of interest in developing inferential tools for RL and other online learning algorithms such as bandits and SGD. \cite{li2019sgd} proposed a statistical inference method for M-estimation problems based on fixed step-size SGD. \cite{chen20sgd} derived two kinds of estimators for the asymptotic variance of the averaged SGD parameter estimate. \cite{fang19} proposed an online estimator for SGD based on randomly perturbing the parameter estimates at each time step. \cite{chen2020statistical} developed an SGD-based algorithm for online decision making, derived an inverse probability weighted value estimator to estimate the optimal value, and proposed plug-in estimators to estimate the variance of the parameters. While all of these methods are well suited to cases where the data is generated by an i.i.d. sampling procedure, they are unable to account for the underlying dependence structure of the observations generated by RL algorithms.

A closely related field of study is that of dynamic treatment regimes (DTR). It generalizes personalized medicine to time-varying treatment settings where treatments are sequentially adapted to a patient's temporal state. There are similarities to the online learning methods studied here, and recent works have explored the application of RL algorithms such as Q-learning to this domain \citep{chakraborty2014}, with an emphasis on quantifying the statistical uncertainty associated with their estimators. \cite{ertefaie2018constructing} develop an estimation procedure for the optimal DTR over an indefinite time period and derive associated large-sample results. \cite{luckett2019estimating} proposed a new RL method for estimating an optimal treatment regime applicable to the mobile health domain with an infinite time horizon, and established the consistency and asymptotic normality of their estimators under relevant assumptions. These works focus on the estimation of the optimal policy, while ours deals with the inference of the value function for a given policy.

Within the RL and bandit settings, there are a number of recent works that consider the problem of uncertainty quantification of the policy's parameter estimates or value function, focusing mainly on the off-policy setting. \cite{zhang2021statistical} showed how M-estimation can be extended to provide inferential methods for data collected with bandit algorithms. \cite{kuzborskij21a} proposed a method for confidence interval estimation based on the Efron-Stein tail inequality within the off-policy contextual bandit setting. Both these methods apply to the bandit setting, where there is no sequential dependence in state transitions. Within the off-policy RL setting, \cite{dai2020coindice} proposed a method for computing confidence intervals for a target policy's value based on an optimization formulation of the value estimation problem. \cite{jiang2020minimax} derive minimax value intervals for off-policy evaluation that satisfy a certain double robustness criterion. However, both papers assume a generative model, where observation tuples are sampled independently from the stationary distribution of the underlying Markov decision process. \cite{shi2021statistical} proposed an inference method for the state-action value (Q) function via sieve methods to approximate the Q-function. This is an offline method that directly computes the value estimates using batch updates. On the other hand, ours is a fully online inference method.

A number of recent works in the RL literature \citep{bhandari18, srikant19, xu20, kaledin2020finite} have focused on establishing finite sample bounds for the value estimates in TD learning and related algorithms. While these methods provide tight non-asymptotic bounds for the value function under standard assumptions, they do not allow for statistical inference of the estimates. By contrast, the distributional consistency guaranteed by our method allows for the construction of asymptotically exact confidence intervals for the value estimates.

Finally, our online bootstrap algorithm is related to a few recent works that have applied bootstrapping to the bandit and RL settings in various contexts. \cite{wang2020residual} proposed a perturbation-based bootstrap exploration method in the bandit setting. \cite{hao2020ucb} utilized multiplier bootstrap to estimate the upper confidence bound for exploration in the bandit settings. Within the RL setting, \cite{white10} used the moving block bootstrap method to compute confidence intervals for value estimates in continuous Markov decision processes. \cite{hanna17} presented two model-based bootstrap methods to compute confidence bounds for off-policy value estimates. \cite{hao2021bootstrapping} proposed a subsampled bootstrap method for the statistical inference of value estimates computed using the fitted Q-evaluation algorithm. All of these methods require access to the entire dataset in order to carry out the resampling procedure and are therefore only applicable to the batch RL setting. In contrast, the method proposed here computes the bootstrap estimates in an online manner and does not require storage of past observations. 

The rest of this paper is organized as follows: In Section \ref{sec:background}, we introduce the linear stochastic approximation algorithm and provide some relevant background on the RL algorithms. In Section \ref{sec:algo}, we present the online bootstrap algorithm. In Section \ref{sec:theory}, we discuss our theoretical results. In Section \ref{sec:exp}, we present some numerical simulations that demonstrate the efficacy of the algorithm in various settings, including on-policy TD learning in the FrozenLake RL environment, deep-Q learning in the Atari Pong RL environment, and off-policy GTD learning in a simulated MDP setting and a real healthcare setting. Finally, in Section \ref{sec:discussion}, we summarize our work and discuss some interesting future directions.

\section{Background}
\label{sec:background}
\subsection{Linear Stochastic Approximation Under Markov Noise}
\label{sec:lsa}

Stochastic approximation is a classic algorithm with a long history in optimization \citep{robbins51sa}. In its linear form, the algorithm is designed to solve the equation $\Abar \theta = \bbar$, where $\Abar \in \R^{d \times d}$ and $\bbar \in \R^d$ are unknown deterministic quantities, and $\theta \in \Theta \subseteq \R^d$ is the parameter of interest. In the present setting, we are given a sequence of observations of the form $\{(\Atilde(X_{t}), \btilde(X_t))\}_{t\geq 1}$, where $\{X_t\}_{t \geq 1}$ is an ergodic Markov chain with state space $\Xcal$ and stationary distribution $\mu$, and $\Atilde: \Xcal \to \R^{d \times d}$ and $\btilde: \Xcal \to \R^d$ are matrix and vector-valued functions defined on the state space $\Xcal$, whose expectations under the stationary distribution $\mu$ are $\Abar$ and $\bbar$, respectively.

The update step for this algorithm is given by
\begin{align}
    \theta_{t+1} = \theta_t + \alpha_{t+1}(\Atilde(X_{t+1})\theta_t - \btilde(X_{t+1})), \label{eq:lsa1}
\end{align}
where $\theta_t \in \Theta$ is the stochastic approximation iterate i.e., the current estimate of $\theta$, and $\{\alpha_t\}_{t \geq 1}$ is a sequence of polynomially decaying step-sizes, \textcolor{black}{i.e., $\alpha_t = \alpha_0 / t^\eta$, for some $\alpha_0 > 0$ and learning rate $\eta \in \left(\frac{1}{2},1 \right)$.}

Our algorithm applies not to the iterate $\theta_t$ itself, but to the averaged iterate $\thetabar_t = \frac{1}{t }\sum_{i=1}^t \theta_i$. This averaging scheme is referred to as Polyak-Ruppert averaging, after \cite{Ruppert1988EfficientEF} and \cite{polyak92}, who established the asymptotic normality of $\thetabar_t$ for strongly convex objective functions under Martingale noise.

\subsection{Policy Evaluation In Reinforcement Learning}
\label{sec:rl_intro}

\textcolor{black}{In this section, we briefly review some background theory on Markov reward processes and policy evaluation, and describe the temporal difference (TD) learning and Gradient TD (GTD) learning algorithms as instances of the linear stochastic approximation algorithm with Markov noise in (\ref{eq:lsa1}). In addition to these, recent studies \citep{mou2021optimal, durmus2021stability, chen2021lyapunov} show that other TD variates like n-step TD and TD($\lambda$) are also special cases of Markovian linear stochastic approximation algorithms, which enables their studies on the finite-sample convergence guarantees of these RL algorithms. Our setting can be viewed an a complementary to the martingale noise setting considered in offline policy evaluation problems \citep{luckett2019estimating, shi2021statistical}. These methods do not allow for an online update but instead consider a batch update or an offline update. }


\subsubsection*{Markov Reward Processes}
A Markov Decision Process (MDP) is denoted as $\Mcal = (\mathcal{S}, \mathcal{A}, \mathcal{P}, \mathcal{R}, \gamma)$, where $\mathcal{S}$ is the state space, $\mathcal{A}$ is a finite set of actions, $\Pcal$ is the transition kernel, $\Rcal$ is the reward function, and $\gamma \in (0,1)$ is a discount factor. 

A stationary policy $\pi$ maps a state $s \in \mathcal{S}$ to a probability space via the distribution $\pi(\cdot |s)$. At time step $t$, suppose the learner at state $s_{t} \in \mathcal{S}$, following the policy $\pi$, takes the action $a_{t} \in \mathcal{A}$ with probability $\pi(a_{t}|s_{t})$. Then the transition kernel $\Pcal(s_{t+1}|s_t, a_{t})$ determines the probability of being in the next state $s_{t+1} \in \mathcal{S}$ at the next time step, and the reward $r_{t+1} = \mathcal{R}(s_{t}, a_{t}, s_{t+1})$, assumed to be bounded by $r_{\max}$, is obtained. 

The stationary policy $\pi$ and the MDP together induce a Markov Reward Process (MRP) $\Mcal^\pi = (\Mcal, \pi)$, with transition kernel $\Pcal^{\pi}(s'|s) = \sum_{a \in \mathcal{A}} \Pcal(s' | s, a) \pi(a | s).$ Similarly, the expected reward function of the MRP is given by $$\Rcal^\pi(s) = \sum_{a \in \Acal} \pi(a | s) \sum_{s' \in \Scal} \Pcal(s' | s, a) \Rcal(s, a, s').$$ The ergodicity of the Markov chain ensures the existence of a \textcolor{black}{stationary state distribution $\mu_\pi$ for the MRP over the state space under the stationary policy $\pi$.}

\subsubsection*{Value Functions}
The value function associated with a policy $\pi$, denoted as $V^\pi: \Scal \to \R$, is the discounted sum of expected rewards from starting at a state and following policy $\pi$: $$V^\pi(s) = \E \left[ \sum_{t=0}^\infty \gamma^t \Rcal^\pi(s_t) | s_0 = s \right], \quad s \in \Scal,$$ where the expectation is taken over the set of trajectories generated according to the transition kernel $\Pcal^\pi$.

The value function is the unique solution to the Bellman equation
\begin{align}
V^\pi(s) = \Rcal^\pi(s) + \gamma \E_{s' \sim \Pcal^\pi(\cdot|s)} \left[ v^\pi(s') \right], \quad s \in \Scal. \label{eq:bellman}
\end{align}

We define the Bellman operator on the space of value functions as $$T^\pi V(s) = \Rcal^\pi(s) + \gamma \E_{s' \sim \Pcal^\pi(\cdot|s)}[V(s')],$$ for any value function $V: \Scal \to \R$. Then $V^\pi$ is the unique fixed point of the operator $T^\pi$.

\subsubsection*{TD Learning With Linear Function Approximation}
TD learning is widely used for the estimation of the value function for a given policy. The classic version of this algorithm, now known as tabular TD learning \citep{sutton88}, attempts to compute value estimates for every state in the state space. In modern RL applications, such an approach becomes infeasible, as real-world problems often involve very large state spaces. In those cases, a natural approach is to approximate the value function as $V_\theta(s) = \sum_{i=1}^d \varphi_i(s) \theta_i = \phi(s)\Tra \theta$, where $\{\varphi_i: \Scal \to \R\}_{i=1}^d$ is a set of basis functions, $\phi(s) = [\varphi_1(s), \ldots, \varphi_d(s)]\Tra$, and $\theta \in \R^{d}$ denotes the parameter. Then, given a sequence of observations of the form $\{(s_t, r_{t+1}, s_{t+1}) \}_{t \geq 0}$, the linear TD update is given by
\begin{align}
    \theta_{t+1} \leftarrow \theta_t + \alpha_{t+1}\left((\phi(s_t) - \gamma \phi(s_{t+1})\Tra \theta_t - r_{t+1} \right)\phi(s_t) \label{eq:td_update}
\end{align}

When the state space $\Scal$ is finite, the states may be enumerated as $\{1, \ldots, |\Scal|\}$, with the integer $i$ corresponding to $s^{(i)}$, the $i$th state in $\Scal$. We can then express the transition kernel $\Pcal^\pi$ as the matrix $P^\pi \in \R^{|\Scal| \times |\Scal|}$, with $P^\pi_{i,j} = \Pcal^\pi(s^{(j)} | s^{(i)})$, and the expected rewards at each state as the vector $r^\pi \in \R^{|\Scal|}$, with $r^\pi_i = \Rcal^\pi(s^{(i)})$.

Using this formulation, linear TD learning may be seen as an instance of the linear stochastic approximation update (\ref{eq:lsa1}), solving the linear equation $\Abar \theta = \bbar$, with $$\Abar = \Phi\Tra \Xi (I - \gamma P^\pi) \Phi, \quad \text{and} \quad \bbar = \Phi\Tra \Xi r^\pi,$$ where $\Phi \in \R^{|\Scal| \times d}$ denotes the feature matrix, containing the features $\phi(s^{(i)}), 1 \leq i \leq |\Scal|$ in its rows, and $\Xi \in \R^{|\Scal| \times |\Scal|}$ is a diagonal matrix with elements corresponding to the entries of the stationary distribution $\mu_\pi$.

For samples $X_{t} = (s_{t}, r_{t+1}, s_{t+1})$ generated sequentially from the MRP $\Mcal^\pi$, we can run linear stochastic approximation using the quantities $$\Atilde(X_{t}) = \phi(s_t)\left(\phi(s_t) - \gamma \phi(s_{t+1}) \right)\Tra, \quad \text{and} \quad \btilde(X_{t}) = r_{t+1}\phi(s_{t}).$$ 

\subsubsection*{GTD Learning For Off-Policy Evaluation}
In off-policy evaluation, the goal is to estimate the value of a target policy $\pi$, given a set of observations generated by a behavior policy $\pi_b$. This is an important problem in RL, as it enables us to evaluate several policies using data generated by a different behavior policy.

Unlike the on-policy setting, the traditional TD learning is no longer feasible due to its convergence issue in the off-policy setting \citep{sutton2018reinforcement}. On the other hand, Gradient TD (GTD) algorithms \citep{sutton08,sutton09} are guaranteed to converge even in the off-policy setting. 

The algorithm uses a form of importance sampling to correct for the discrepancy between the target and behavior policy. This is done by scaling the updates by an importance sampling ratio $\rho_t = \frac{\pi(a_t | s_t)}{\pi_b(a_t | s_t)}$.

There are two variants of this algorithm, GTD1 and GTD2, which seek to minimize the Norm of the Expected TD Update (NEU), and the Mean-Square Projected Bellman Error (MSPBE), respectively. These loss functions have a similar structure, and can be unified as
\begin{align*}
J(\theta) = \| \Phi\Tra \Xi (r^\pi + \gamma P^\pi \hat{v}_\theta - \hat{v}_\theta) \|^2_{M^{-1}},
\end{align*}
where $M \in \R^{d \times d}$ is the identity under NEU, and $M = \Phi\Tra \Xi \Phi$ under MSPBE.

In order to minimize this loss function, a pseudo-stochastic gradient method was proposed, involving two simultaneous updates:
\begin{align*}
    y_{t+1} & = y_t + \alpha_{t+1}(b_t + A_t \theta_t - M_t y_t),  \\
    \theta_{t+1} & = \theta_t + \alpha_{t+1} A_t^T y_t,
\end{align*}
where $b_t = \rho_t r_{t+1} \phi(s_t)$ and $A_t = \rho_t \phi(s_t) \left( \phi(s_t) - \gamma \phi(s_{t+1}) \right)\Tra$ are unbiased estimators for $b = \Phi\Tra \Xi r^\pi$ and $A = \Phi\Tra \Xi (I - \gamma P^\pi) \Phi$, respectively. Similarly, $M_t$ is an unbiased estimator for $M$, with $M_t = I_d$ under NEU and $M_t = \phi(s_t) \phi(s_t)\Tra$ under MSPBE. 

These two steps can be combined into a single linear stochastic approximation update solving the linear equation $\Abar \Theta = \bbar$, where
$$\Abar = \begin{pmatrix} 0 & -A\Tra \\ A & M \end{pmatrix}, \quad \bbar = \begin{pmatrix} 0 \\ b \end{pmatrix},\quad \text{and} \quad \Theta = \begin{pmatrix} \theta \\ y \end{pmatrix}.$$

As in case of TD learning with linear function approximation, for samples generated sequentially under the behavior policy $\pi_b$, we denote the observed tuples as $X_t = (s_t, r_{t+1}, s_{t+1})$. Then we can run linear stochastic approximation using the quantities
$$\Atilde(X_t) = \begin{pmatrix} 0 & -A_t\Tra \\ A_t & M_t \end{pmatrix}, \quad \btilde(X_t) = \begin{pmatrix} 0 \\ b_t \end{pmatrix}, \quad \text{and} \quad \Theta_t = \begin{pmatrix} \theta_t \\ y_t \end{pmatrix}.$$

\section{Method}
\label{sec:algo}
\subsection{Online Bootstrap Algorithm}
Given a dataset $\{\Atilde(X_t), \btilde(X_t)\}_{t\geq1}$ of sequentially generated observations, the Polyak-Ruppert averaged iterate estimates the parameter $\theta_*$ as $\thetabar_t = \frac{1}{t} \sum_{i=1}^t \theta_i$, where $\theta_i$ is defined in $(\ref{eq:lsa1})$. Under the assumptions listed in Section \ref{sec:theory}, $\thetabar_t$ is a consistent estimator of $\theta_*$, and its distribution is asymptotically Gaussian with mean $\theta_*$ and a certain covariance matrix. The estimation of this asymptotic covariance is crucial for statistical inference, but there is presently no straightforward way to estimate it in the presence of Markov noise. This motivates the development of the online bootstrap method.

The online bootstrap method is based on the following perturbed stochastic approximation update, which is performed in parallel to the main update (\ref{eq:lsa1}):
\begin{align}
    \thetahat_{t+1} = \thetahat_t + \alpha_{t+1} W_{t+1} (\tilde{A}(X_{t+1}) \thetahat_t - \tilde{b}(X_{t+1})), \label{eq:plsa1}
\end{align}
where $\{W_t\}_{t \geq 1}$ is a bounded sequence of i.i.d. random variables with mean 1 and variance 1, with $W_1 < W_{\max}$ a.s. for some finite constant $W_{\max} > 0$. The $\hat{\cdot}$ notation is used here to distinguish the perturbed iterates from those of the standard update (\ref{eq:lsa1}). Let $\bar{\thetahat}_t$ denote the averaged iterate of the sequence $\{\thetahat_t\}_{t \geq 0}$. In Theorem \ref{thm:plsa3} of the Section \ref{sec:theory}, we show that the distributions of $\sqrt{t}(\thetabar_t - \theta_*)$ and $\sqrt{t}(\bar{\thetahat}_t - \thetabar_t)$ are asymptotically equivalent. This is a fundamental result for the validity of our online bootstrap algorithm, as it enables us to conduct inference on the former distribution by using the latter as a proxy. The latter distribution may be approximated by bootstrapping $B$ samples of $\bar{\thetahat}_t - \thetabar_t$. For each $b = 1,\ldots, B$, at time step $t+1$, we update the perturbed SGD iterates $\thetahat^{(b)}_t$ as follows:
\begin{align*}
    \thetahat^b_{t+1} & = \thetahat^b_{t} + \alpha_{t+1} W^b_{t+1}(\Atilde(X_{t+1})\thetahat^b_t - \btilde(X_{t+1})), \\
    \bar{\thetahat}^b_{t+1} & = \frac{1}{t+1}\sum_{i=1}^{t+1} \thetahat^b_i,
\end{align*}
where $W^b_t$ are i.i.d. random variates with mean one and variance one. The updates can be performed in a fully online manner, as they only rely on the latest available data point $X_{t+1}$. Furthermore, since all $B$ trajectories of perturbed iterates depend on a single trajectory of the Markov chain $\{X_t\}$, the iterates can be updated in parallel. Intuitively, the bootstrap method proposed here circumvents the issue of higher-order dependence in the SA iterates by enabling us to perform statistical inference using the cross-section of perturbed iterates generated at each time step, as opposed to using the highly dependent iterates generated by a single trajectory. Algorithm \ref{alg:lsa_bootstrap} presents the entire online update scheme. 

\begin{algorithm}[h]
\DontPrintSemicolon
\SetKwInOut{Input}{Input}
\SetKwInOut{Output}{Output}
\Input{Number of bootstrap samples $B$, Initial step size $\alpha_0 > 0$, Learning rate $\eta \in (\frac{1}{2},1)$, Initial estimates $\theta_0 = \thetahat_0^b, b = 1,\ldots,B$.}
\For{$t=0,1,2,\ldots$}{
    Observe $\Atilde(X_{t+1}), \btilde(X_{t+1})$.\;
    Compute $\alpha_{t+1} = \alpha_0 \eta^{-(t+1)}$.\;
    Update $\theta_{t+1} \leftarrow \theta_t + \alpha_{t+1} (\Atilde(X_{t+1})\theta_t - \btilde(X_{t+1}) ) $.\;
    Update $\thetabar_{t+1} \leftarrow \frac{1}{t+1}(t \thetabar_t + \theta_{t+1})$.\;
    \For{$b = 1,2,\ldots,B$}{
        Update $\thetahat^b_{t+1} \leftarrow \thetahat^b_t + \alpha_{t+1} W_{t+1}^b (\Atilde(X_{t+1})\thetahat^b_t - \btilde(X_{t+1}) )$.\;
        Update $\bar{\thetahat}^b_{t+1} \leftarrow \frac{1}{t+1}(t \bar{\thetahat}^b_t + \thetahat^b_{t+1})$.
    }
}
\Output{Bootstrap estimates $\left\{\bar{\thetahat}_{t+1}^b \right\}_{b=1}^B$}

\caption{Online Bootstrap For Linear Stochastic Approximation}
\label{alg:lsa_bootstrap}
\end{algorithm}

\subsection{Constructing Confidence Intervals}
We use two approaches to construct confidence intervals using the bootstrap empirical distribution - the quantile and the standard error estimators. As the names suggest, these are based on estimating the quantiles and standard errors of the bootstrap errors. Let $q_\delta$ denote the $\delta$th quantile of the empirical bootstrap distribution $\left\{\bar{\thetahat}^{(b)}_t - \thetabar_t\right\}_{b=1}^B$. Then the $(1-\alpha)$ quantile-based confidence interval for $\theta$ is given by $(\thetabar_t + q_{\alpha/2}, \thetabar_t + q_{1 - \alpha/2})$. 

Similarly, let $\hat{\Sigma}$ denote the sample covariance matrix of the empirical distribution, and let $\hat{\sigma} = \sqrt{\text{diag}(\hat{\Sigma} )}$. Then the $(1-\alpha)$ standard error-based confidence interval for $\theta$ is given by $(\thetabar_t + z_{\alpha/2}\hat{\sigma}, \thetabar_t + z_{1-\alpha/2} \hat{\sigma})$, where $z_\alpha$ denotes the $\alpha$th quantile of the standard normal distribution. The validity of the SE confidence interval relies on the consistency of the second moment of $\thetabar_t$. Moment consistency is directly implied by the distributional consistency result of Theorem \ref{thm:plsa3} under slightly stronger conditions \citep{cheng15}.

\color{black}
In the task of policy evaluation in RL, we are interested in constructing confidence intervals for the value function. Under linear function approximation, the value estimate corresponding to the policy $\pi$, given the averaged iterate $\thetabar_t$, is $v_{\thetabar_t}(s) = \phi(s)\Tra \thetabar_t$, where $\phi(s)$ denotes the feature mapping for the state $s \in \Scal$. More generally, we are interested in estimating the value associated with a reference distribution $\nu$ over the state space $\Scal$. In this case, the value estimate for the policy $\pi$ is given by
\begin{align*}
    V^\pi_{\thetabar_t}(\nu) = \int_{s \in \Scal} \phi(s)\Tra {\thetabar_t} \nu(ds).
\end{align*}
Confidence intervals for $V^\pi_{\thetabar_t}(\nu)$ can be constructed using the quantile or standard error estimators. Let $q_\delta$ denote the $\delta$th quantile of the bootstrapped value estimates $\left\{V^\pi_{\bar{\thetahat}_t^{(b)}}(\nu) - V^\pi_{\thetabar_t}(\nu) \right\}_{b=1}^B$. Then the $(1-\alpha)$ quantile-based confidence interval for the value estimate is given by $$\left(V^\pi_{\thetabar_t}(\nu) + q_{\alpha/2}, V^\pi_{\thetabar_t}(\nu) + q_{1 - \alpha/2}\right).$$ 

Similarly, let 
\begin{align*}
    \tilde{\sigma}(\nu) = \left( \int_{s \in \Scal} \phi(s) \nu(ds) \right)\Tra \hat{\Sigma} \left( \int_{s \in \Scal} \phi(s) \nu(ds) \right)
\end{align*}
denote the sample covariance matrix of the bootstrapped value estimates. Then the $(1-\alpha)$ standard error-based confidence interval for the value estimate is given by $$\left(V^\pi_{\thetabar_t}(\nu) + z_{\alpha/2}\tilde{\sigma}(\nu), V^\pi_{\thetabar_t}(\nu) + z_{1-\alpha/2} \tilde{\sigma}(\nu)\right).$$

\color{black}
\section{Distributional Consistency Result}
\label{sec:theory}
\subsection{Asymptotic Theory For Linear Stochastic Approximation Under Markov Noise}
\label{sec:theory_lsa}
In this section, we list our assumptions and state some relevant results pertaining to the asymptotic theory of linear stochastic approximation.

\color{black}
\begin{assumption} \label{asu1} 
    The Markov chain $\{X_t\}$ is uniformly ergodic, with transition kernel $\Pcal$ and unique stationary distribution $\mu$.
\end{assumption}

Assumption \ref{asu1} is standard in the RL literature for the policy evaluation setting, especially in recent works on the finite-sample analysis of policy evaluation algorithms \citep{bhandari18, srikant19, xu20}. It always holds for irreducible, aperiodic Markov chains \citep{levin2017markov}.

As a direct consequence of \ref{asu1}, there exist constants $M > 0$ and $\kappa \in (0,1)$ such that 
\begin{align}
    \sup_{x \in \Xcal} \left\|\Pcal^t(x,\cdot) - \mu \right\| \leq M \kappa^t, \label{eq:mixing}
\end{align}
where $\Pcal^t(x,\cdot)$ denotes the $t$-step transition kernel, starting from state $x$ (see e.g. \cite{meyn09, douc2018}). Equation (\ref{eq:mixing}) characterizes the rate at which the Markov chain $\{X_t\}$ approaches its stationary distribution $\mu$, when starting from an arbitrary initial distribution.

\color{black}
Next, we define conditions on the noisy observations:

\begin{assumption} \label{asu2}
    \subassumption \label{asu2i} There exists a matrix $\Abar$ and a vector $\bbar$ such that 
    \begin{align*}
        \Abar = \lim_{t \to \infty}\E[\Atilde(X_t)], \quad \text{and} \quad \bbar = \lim_{t \to \infty}\E[\btilde(X_t)].
    \end{align*}
    \subassumption \label{asu2ii} The matrix $\Abar$ is full rank and Hurwitz, i.e., all its eigenvalues have strictly negative real parts. \\
    \subassumption \label{asu2iii} There exist constants $A_{\max}$ and $b_{\max}$ such that
    \begin{align*}
        \sup_{x \in \Xcal}\left\|\Atilde(x) \right\|_F \leq A_{\max}, \quad \sup_{x \in \Xcal}\left\|\btilde(x) \right\|_2 \leq b_{\max}.
    \end{align*}
\end{assumption}

The conditions listed in \ref{asu2} are imposed in order to ensure convergence of the sequence of iterates $\{\theta_t\}$. \ref{asu2i} is self-explanatory, while \ref{asu2ii} is a standard assumption in the stochastic approximation literature that ensures the stability of the algorithm \citep{polyak92, srikant19, chen20sgd}. It is generally considered a reasonable assumption in the RL setting, both for TD learning \citep{bhandari18, hu19} and for GTD learning \citep{gupta19}. \ref{asu2iii} controls the behavior of the Markov noise. In the RL setting, it holds whenever the feature maps $\phi$ and the reward function $\Rcal$ are bounded \citep{sutton2018reinforcement}. By \ref{asu2i} and \ref{asu2iii}, we also have that $\left\|\Abar \right\|_F \leq A_{\max}$ and $\left\|\bbar \right\|_2 \leq b_{\max}$.

By (A2), there exists a unique solution $\theta_* \in \Theta$ to the linear equation $\Abar \theta = \bbar$. Furthermore, we can now write (\ref{eq:lsa1}) as
\begin{align}
    \theta_{t+1} = \theta_t + \alpha_{t+1} (\Abar \theta_t - \bbar) + \alpha_{t+1} \epsilon_{t+1}, \label{eq:lsa2}
\end{align}
where $\epsilon_{t+1} = (\Atilde(X_{t+1}) - \Abar)\theta_t - (\btilde(X_{t+1}) - \bbar)$ is a residual noise term. 

Our final assumption has to do with the step sizes $\{\alpha_t\}$:

\begin{assumption} \label{asu3}
The step sizes are of the form $\alpha_t = \alpha_0 / t^\eta$, $t \geq 1$, where $\alpha_0 > 0$ and the learning rate $\eta \in (\frac{1}{2}, 1)$.
\end{assumption}

Polynomially decaying step sizes are standard in the case of iterate averaging \citep{polyak92}.

\color{black}
Under these conditions, we have the following almost sure rate of convergence result for the update (\ref{eq:lsa2}):

\begin{proposition}
\label{thm:lsa_as}
Suppose that (A1)-(A3) hold. Let $\eta \in (1/2,1)$ be defined as in (A3), and let $\gamma \in (0, \eta-1/2)$. Then the iterates of update (\ref{eq:lsa2}) satisfy $\left\| \theta_t - \theta_* \right\|_2 = o(t^{-\gamma})$, a.s.
\end{proposition}

This result establishes the consistency of the LSA iterate $\theta_t$ in the Markov noise setting. Additionally, it ensures that the iterates are bounded within a compact set without the need for a projection scheme. A proof of the result is provided in the supplementary section. 

\color{black}
By Lemma A.5 of \cite{liang2010}, we can split the noise term in (\ref{eq:lsa2}) into three parts as
\begin{align}
    \epsilon_t = e_t + \nu_t + \zeta_t, \label{eq:markov_noise}
\end{align}
where $e_t$ is a martingale difference sequence, i.e., $\E[e_t | \Fcal_{t-1}] = 0$, where $\Fcal_t$ is the natural filtration associated with the Markov chain $\{X_t\}$, while $\nu_t$ and $\zeta_t$ are decaying residual noise terms.

\begin{proposition}
\label{thm:lsa_clt}
Assume conditions (A1)-(A3) hold. Then $$\sqrt{t}(\bar{\theta}_t - \theta_*) \implies \Ncal(0, \Abar^{-1} Q (\Abar^{-1})\Tra),$$ where $Q = \lim_{t \to \infty}\E[e_t e_t\Tra]$, with $e_t$ defined as in (\ref{eq:markov_noise}).
\end{proposition}

Proposition \ref{thm:lsa_clt} establishes a central limit theorem for the averaged iterate and provides an explicit form for the asymptotic variance. Under i.i.d. noise, the asymptotic variance may be estimated using a plug-in estimator \citep{chen2020statistical}. However, in the present setting, we have $Q = \lim_{t \to \infty}\E[e_t e_t\Tra]$, where $e_t$ is the martingale component of the Markov noise term $\epsilon_t$, as per the decomposition (\ref{eq:markov_noise}). To the best of our knowledge, there is no existing method to estimate $e_t$ by separating it from the other components of $\epsilon_t$. Our online bootstrap algorithm provides an efficient and theoretically sound way to estimate the distribution of $\bar{\theta}_t$. The properties of its key component, the perturbed linear stochastic approximation, are discussed in the following section.

\subsection{Perturbed Linear Stochastic Approximation}
\label{sec:theory_plsa}
We now study the asymptotic behavior of the perturbed linear stochastic approximation update (\ref{eq:plsa1}). As in the standard case, the mean field here is $h(\theta) = \Abar \theta - \bbar$, while the observations are of the form $H(\theta, X_{t+1}) = W_{t+1}(\Atilde(X_{t+1} \theta - \btilde(X_{t+1}))$. By the independence and boundedness of $W_t$, $H(\theta, X_{t+1})$ is an unbiased estimator of $h(\theta)$ under the stationary distribution of $X_t$. 

We may rewrite (\ref{eq:plsa1}) in terms analogous to (\ref{eq:lsa2}), as follows:
\begin{align}
    \thetahat_{t+1} = \thetahat_t + \alpha_{t+1}(\Abar \thetahat_t - \bbar) + \alpha_{t+1}\hat{\epsilon}_{t+1}, \label{eq:plsa2}
\end{align}
where $\hat{\epsilon}_{t+1} = (W_{t+1}\Atilde(X_{t+1}) - \Abar)\thetahat_t - (W_{t+1}\btilde(X_{t+1}) - \bbar)$ is the noise term.

Having established the almost sure rate of convergence for the standard LSA update (\ref{eq:lsa2}), the result may be straightforwardly extended to the perturbed update (\ref{eq:plsa2}):

\color{black}
\begin{proposition}
\label{thm:plsa_as}
Suppose that (A1)-(A3) hold. Let $\eta \in (1/2,1)$ be defined as in (A3), and let $\gamma \in (0, \eta-1/2)$. Then the iterates of update (\ref{eq:plsa2}) satisfy $\left\| \thetahat_t - \theta_* \right\|_2 = o(t^{-\gamma})$, a.s.
\end{proposition}

\color{black}
Next, we establish the theoretical validity of the online bootstrap algorithm of Section \ref{sec:algo}. For this, we need the following lemma:

\begin{lemma}
\label{lem:plsa2}
Assume that (A1)-(A3) hold. Then
\begin{align}
    \sqrt{t} (\bar{\thetahat}_t - \theta_*) = -\frac{1}{\sqrt t} \Abar^{-1} \sum_{i=1}^t W_{i+1}(\Atilde(X_{i+1})\theta_* - \btilde(X_{i+1})) + o_p(1). \label{eq:theorem_2_2a}
\end{align}
\end{lemma}

The proof is provided in the supplementary section. Note that Lemma \ref{lem:plsa2} also holds for the update (\ref{eq:lsa2}), since (\ref{eq:plsa2}) reduces to (\ref{eq:lsa2}) when $W_i \equiv 1$ for all $i$. Hence we have
\begin{align}
    \sqrt{t} (\bar{\theta}_t - \theta_*) = -\frac{1}{\sqrt t} \Abar^{-1} \sum_{i=1}^t (\Atilde(X_{i+1})\theta_* - \btilde(X_{i+1})) + o_p(1). \label{eq:theorem_2_2b}
\end{align}
Subtracting (\ref{eq:theorem_2_2b}) from (\ref{eq:theorem_2_2a}), we get
\begin{align}
    \sqrt{t} (\bar{\thetahat}_t - \bar{\theta}_t) = -\frac{1}{\sqrt t} \Abar^{-1} \sum_{i=1}^t (W_{i+1} - 1)(\Atilde(X_{i+1})\theta_* - \btilde(X_{i+1})) + o_p(1). \label{eq:theorem_2_2c}
\end{align}

The final step is to establish the distributional consistency of the online bootstrap estimator in the Kolmogorov metric. \textcolor{black}{Suppose that the data $\Dcal$ is generated under the probability space $\{ \Xcal, \Acal, \mathbb{P}_\Dcal \}$, while the bootstrap weights $\Wcal = \{W_i\}_{i=1}^t$ are generated under an independent probability space $(\Omega, \Bcal, \mathbb{P}_\Wcal)$. Let $\Pbb_{\Wcal | \Dcal}$  denote the conditional distribution given the observed data $\Dcal$.}

\begin{theorem}
\label{thm:plsa3}
\textcolor{black}{Assume that (A1)-(A3) hold. Then, as $B \to \infty$ and $t \to \infty$, we have}
\begin{align}
\label{eq:kolmogorov}
    \textcolor{black}{\sup_{v \in \R^d}\left| \Pbb_{\Wcal | \Dcal}\left(\sqrt{t}(\bar{\thetahat}_t - \bar{\theta}_t) \leq v\right) - \Pbb_\Dcal \left(\sqrt{t}(\bar{\theta}_t - \theta_*) \leq v \right) \right| \to 0, \textrm{in probability}.}
\end{align}
\end{theorem}

The proof is provided in the supplementary section. We note that the above result applies to the sequence of probability measures $\Pbb_{W | \Dcal}$, where the datasize and the number of bootstrap samples grows to infinity. In practice, there is a bootstrap Monte-Carlo error associated with re-sampling from a finite number of bootstrap samples $B$. By choosing $B$ adequately large, the bootstrap Monte-Carlo error is generally ignored \citep{DasGupta2008}.

Theorem \ref{thm:plsa3} establishes the theoretical foundation for using our online bootstrap estimator for statistical inference. The distributional consistency of the bootstrap estimator in terms of the Kolmogorov metric (\ref{eq:kolmogorov}) enables us to construct asymptotically exact confidence intervals for the estimator $\thetahat$ or functions of the estimator, such as the value estimate in TD learning under linear function approximation.

\section{Experiments}
\label{sec:exp}
In this section, we evaluate the performance of the online bootstrap algorithm through numerical simulations across various settings. We construct two types of confidence intervals using the online bootstrap algorithm - namely, the quantile estimator and the standard error estimator, as defined in the previous section. All confidence intervals were constructed to have a coverage of 95\%. In all cases, we set the number of bootstrap samples $B$ to 200, the learning rate $\eta$ to $3/4$, and the initial parameter $\theta_0$ to the zero vector. \textcolor{black}{The random variates $W_t$ are sampled from the uniform distribution over the interval $(1 - 1/\sqrt{3}, 1 + 1/\sqrt{3})$, so they are bounded and have mean 1, variance 1.}

\textcolor{black}{While our results apply most naturally to the infinite-horizon (continual) setting due to our assumption of ergodicity and the existence of a unique stationary distribution, they can be applied to the finite-horizon (episodic) setting in a straightforward manner. To accomplish this, we simply concatenate the individual episodes to form a trajectory of infinite length, and perform inference on the parameters with respect to this infinite-horizon embedding of the MDP. Similar approach has also been applied in the experiments of other RL literature \citep{dai2020coindice, xu20}. The theoretical validity of such a concatenation procedure has been studied by \cite{bojun2020episodic}.}

The main benchmark we use to measure our algorithm's performance is the vanilla bootstrap, which is an offline method that requires the entire batch of samples for computation. Bootstrap-based inference is highly versatile with regard to the choice of policy evaluation algorithm, and is known to provide better second-order accuracy even when the asymptotic distribution is available \citep{Hall1992bootstrap}. It has already been studied in the RL literature in various contexts \citep{white10, hanna17, hao2021bootstrapping}, and naturally allows for a like-for-like comparison with our method. Therefore, we opt to use the vanilla bootstrap as our primary comparison method. In order to ensure a fair comparison, we use the same number of bootstrap samples, and within each sample, we use the same number of re-sampled observations as the size of the original dataset. 

\subsection{On-Policy Value Inference For FrozenLake RL Environment}
\label{sec:frozenlake}
Next, we consider the Frozenlake environment from OpenAI gym \citep{openai}. Here the RL agent controls the movement of a character in an $8 \times 8$ grid world. The starting point is the first tile of the grid, and the goal is to reach the end tile of the grid. The rest of the tiles are either walkable or absorbing states. A reward of 1 is awarded if the character reaches the target tile, and the reward for any other state transition is 0.

We use linear TD learning to estimate the value function associated with a near-optimal policy trained using Q-learning \citep{sutton2018reinforcement}. For the compared vanilla (offline) bootstrap method, we choose to resample the observations by episodes, rather than by sample transitions, as suggested by \cite{hao2021bootstrapping}, as the sample transitions may fail to capture the sequential dependence in state transitions. 

\begin{figure}[h]
    \centering
    \begin{subfigure}{0.485\textwidth}
        \includegraphics[width=\hsize]{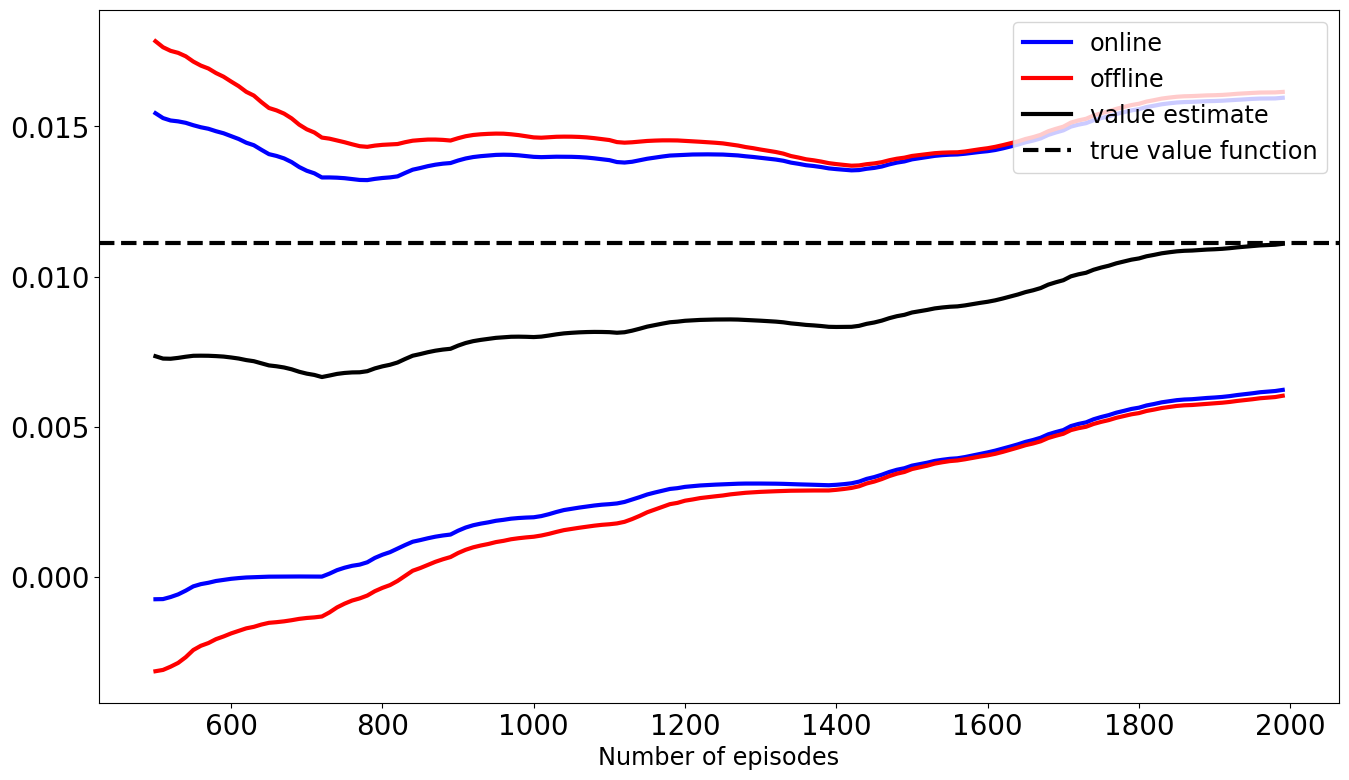}
        \caption{CI Example}
        \label{fig:frozenlake_CI_example}
    \end{subfigure}%
    \hspace{0.02\textwidth}%
    \begin{subfigure}{0.485\textwidth}
        \includegraphics[width=\hsize]{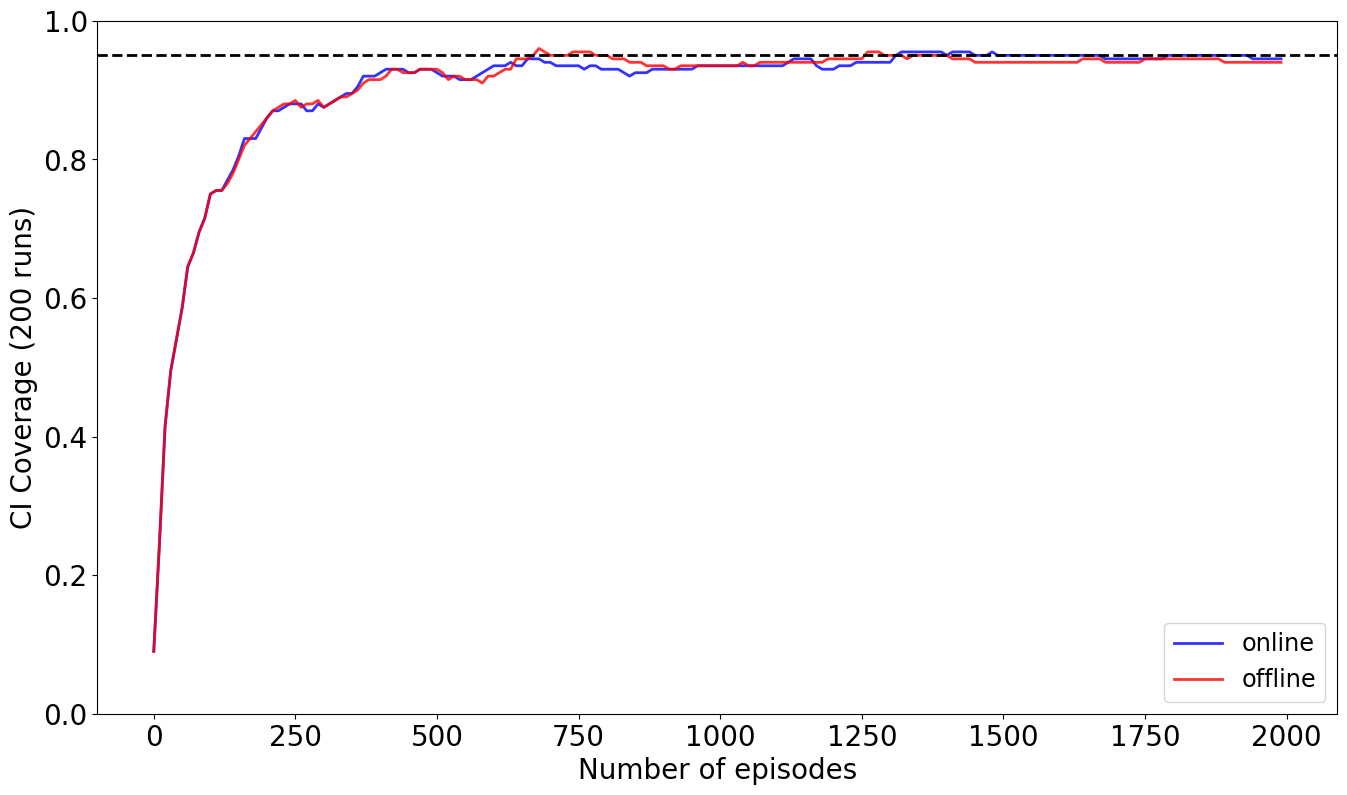}
        \caption{Empirical Coverage Probabilities}
        \label{fig:frozenlake_CI_coverage}
    \end{subfigure}%
    \caption{Statistical inference in the Frozen Lake RL environment: Figure \ref{fig:frozenlake_CI_example} shows an example trajectory for the CIs generated using the online and offline bootstrap methods. The value estimates are for the initial state; the true value for that state is included for comparison. Figure \ref{fig:frozenlake_CI_coverage} shows the empirical coverage probabilities for the value estimates of the initial state, based on 200 repeated experiments using the online and offline bootstrap methods, with 2000 episodes per run.}
    \label{fig:frozenlake1}
\end{figure}

Figure \ref{fig:frozenlake_CI_example} shows an example of the confidence intervals generated for the value estimate of the initial state, using both the online and offline bootstrap methods. The true value function for that state, computed analytically using the transition probability matrix, is included for reference. As in the previous experiment, we also examine the empirical coverage probabilities, in Figure \ref{fig:frozenlake_CI_coverage}. Both methods are seen to achieve the nominal coverage of 95\% within approximately 1000 episodes.

\begin{figure}[h!]
    \centering
    \begin{subfigure}{0.485\textwidth}
        \includegraphics[width=\hsize]{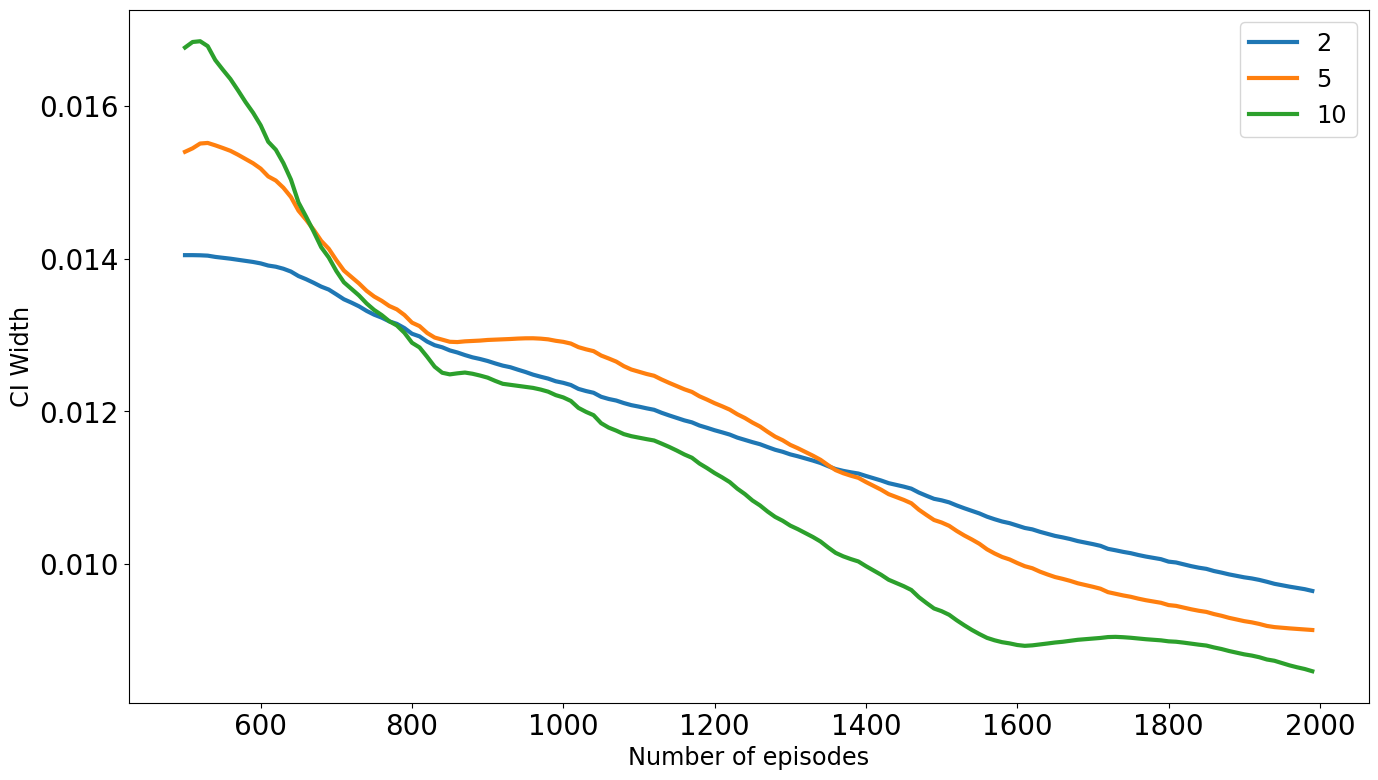}
        \caption{Sensitivity with respect to $\alpha_0$}
        \label{fig:frozenlake_sensitivity_width_alpha0}
    \end{subfigure}%
    \hspace{0.02\textwidth}%
    \begin{subfigure}{0.485\textwidth}
        \includegraphics[width=\hsize]{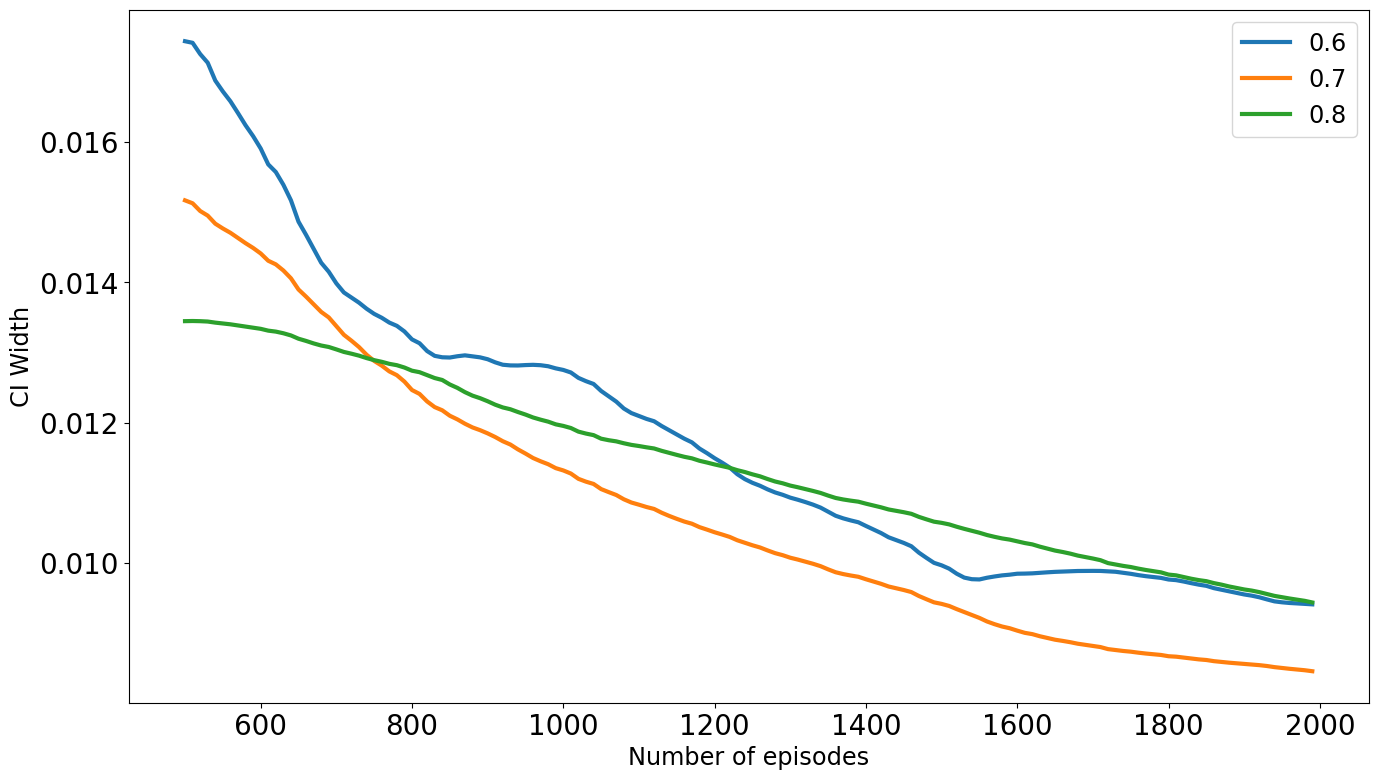}
        \caption{Sensitivity with respect to $\eta$}
        \label{fig:frozenlake_sensitivity_width_eta}
    \end{subfigure}%
    \caption{\textcolor{black}{Statistical inference in the Frozen Lake RL environment: Figure \ref{fig:frozenlake_sensitivity_width_alpha0} shows the sensitivity of the online bootstrap CI widths with respect to the initial step size $\alpha_0$. The legend specifies the values of the initial step sizes used. Similarly, Figure \ref{fig:frozenlake_sensitivity_width_eta} shows the sensitivity of the CI widths with respect to the learning rate $\eta$.}}
    \label{fig:frozenlake_sensitivity_width}
\end{figure}

\begin{figure}[h!]
    \centering
    \begin{subfigure}{0.485\textwidth}
        \includegraphics[width=\hsize]{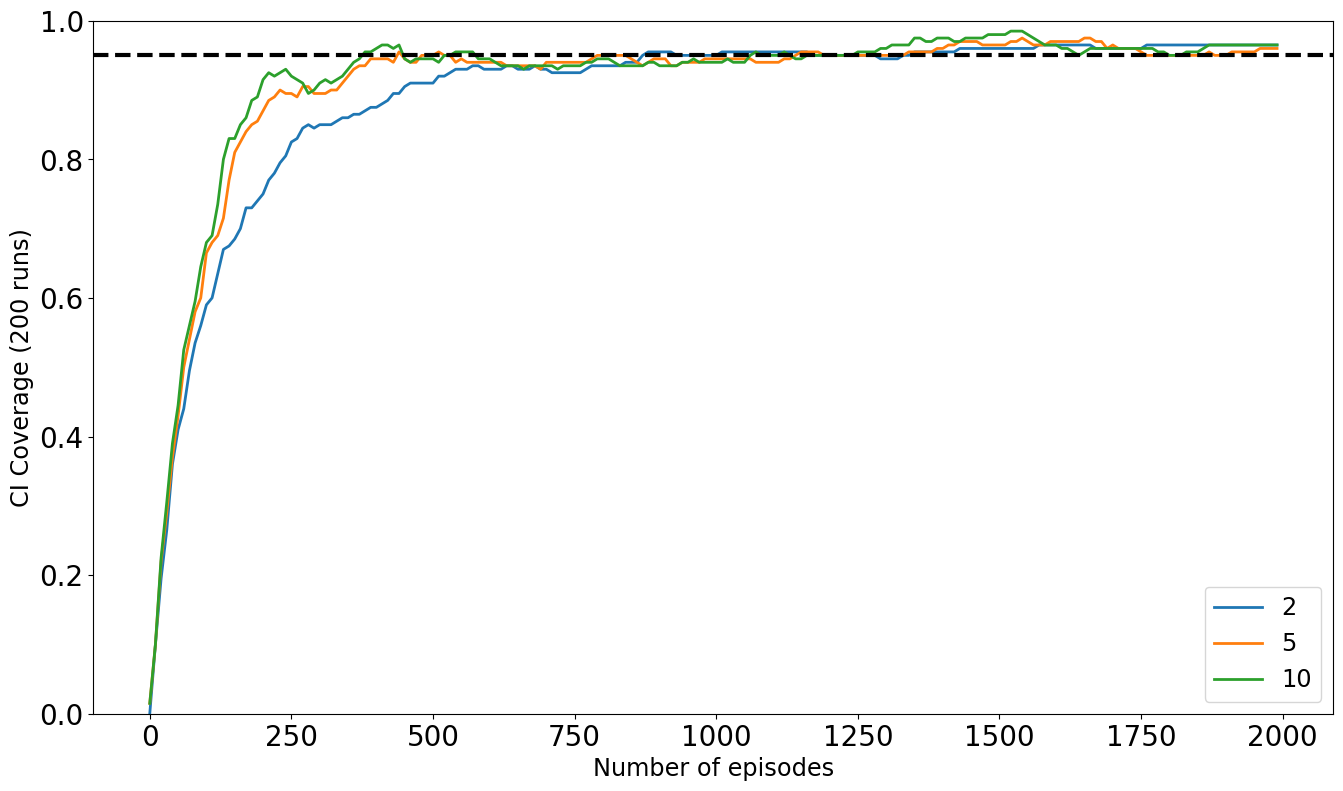}
        \caption{Sensitivity with respect to $\alpha_0$}
        \label{fig:frozenlake_sensitivity_coverage_alpha0}
    \end{subfigure}%
    \hspace{0.02\textwidth}%
    \begin{subfigure}{0.485\textwidth}
        \includegraphics[width=\hsize]{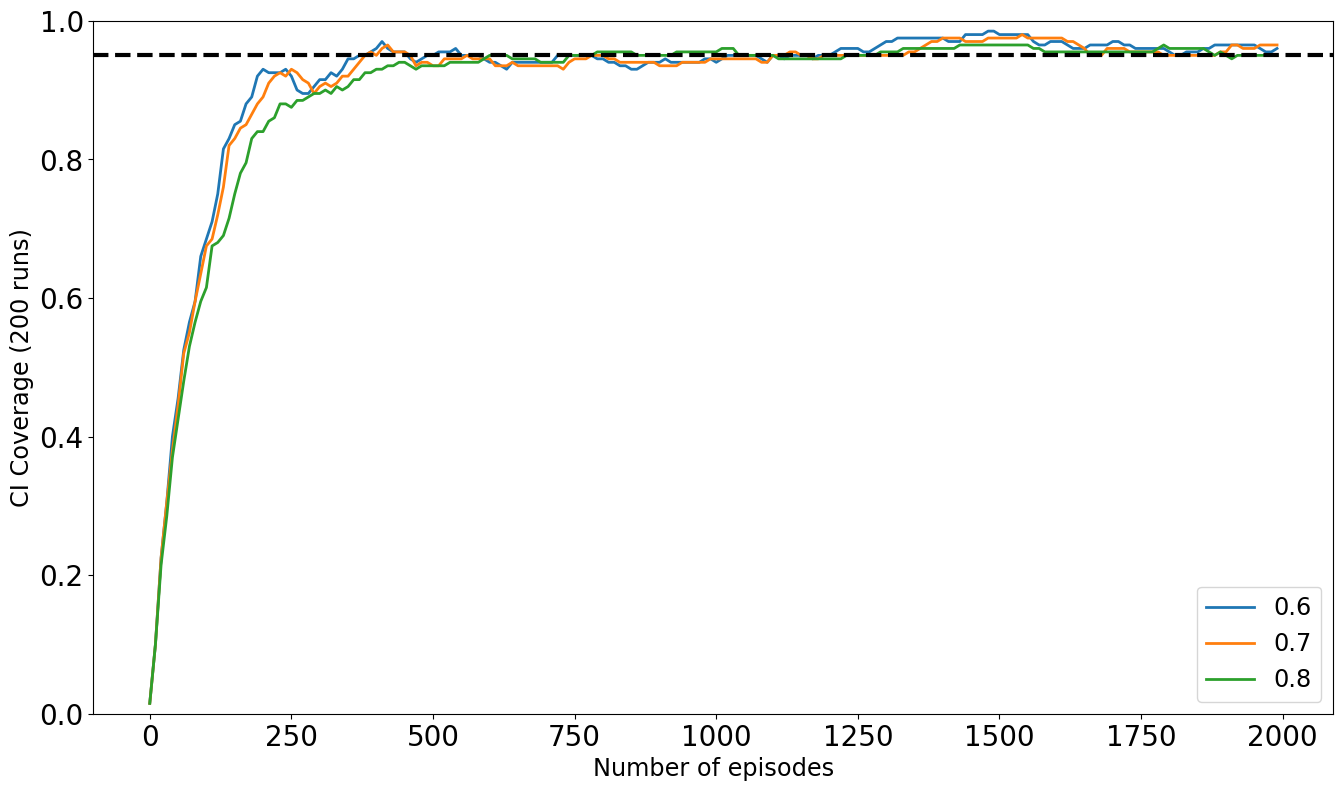}
        \caption{Sensitivity with respect to $\eta$}
        \label{fig:frozenlake_sensitivity_coverage_eta}
    \end{subfigure}%
    \caption{\textcolor{black}{Statistical inference in the Frozen Lake RL environment: Figure \ref{fig:frozenlake_sensitivity_coverage_alpha0} shows the sensitivity of the online bootstrap CI empirical coverage with respect to the initial step size $\alpha_0$. The legend specifies the values of the initial step sizes used. Similarly, Figure \ref{fig:frozenlake_sensitivity_coverage_eta} shows the sensitivity of the empirical coverage with respect to the learning rate $\eta$.}}
    \label{fig:frozenlake_sensitivity_coverage}
\end{figure}

\textcolor{black}{Figure \ref{fig:frozenlake_sensitivity_width} shows the sensitivity of the widths of the CIs generated by the online bootstrap method with respect to the initial step size $\alpha_0$ and the learning rate $\eta$. Similarly, Figure \ref{fig:frozenlake_sensitivity_coverage} shows the sensitivity of the empirical coverage of the generated CIs with respect to the step size parameters. These figures demonstrate that the online bootstrap method is quite robust with respect to the step size, which is one of the only user-defined parameters in the algorithm (alongside the choice of distribution for the random perturbations $W_t$).}

\subsection{On-policy Value Inference For Atari Pong RL Environment}
\begin{figure}[h]
    \centering
    \begin{subfigure}{0.485\textwidth}
        \centering
        \includegraphics[width=0.4\hsize]{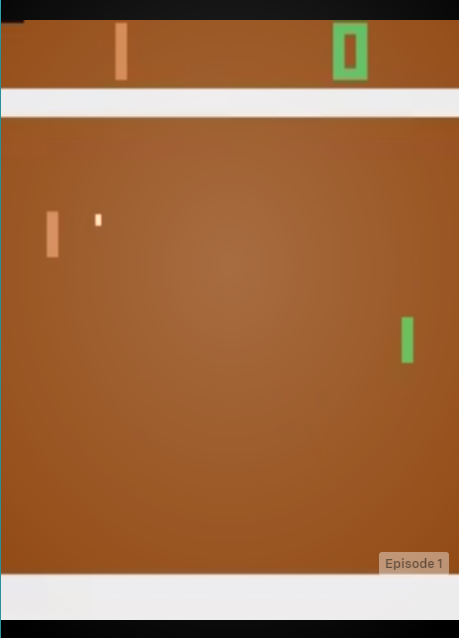}
        \caption{Example Atari Pong game}
        \label{fig:atari_pong_example}
    \end{subfigure}%
    \hspace{0.02\textwidth}%
    \begin{subfigure}{0.485\textwidth}
        \includegraphics[width=\hsize]{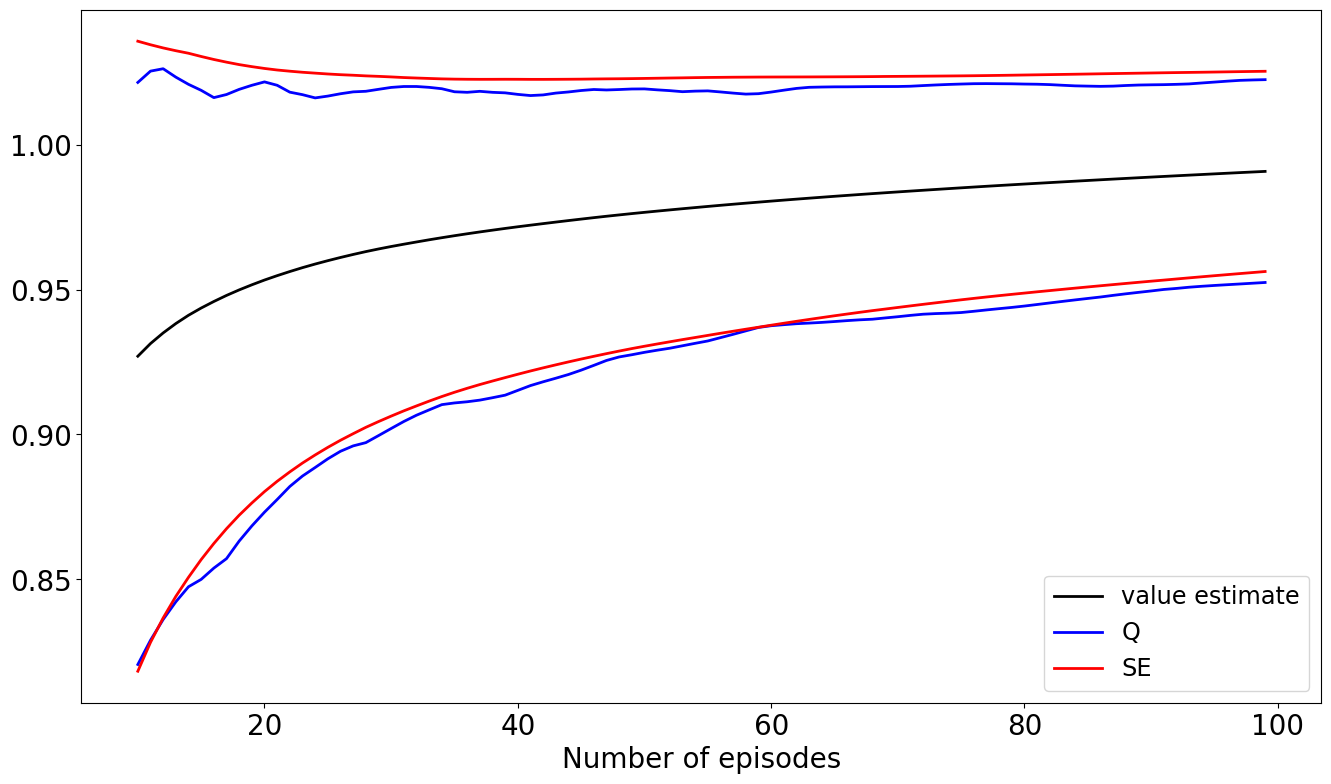}
        \caption{Example trajectory for value estimate and confidence intervals}
        \label{fig:atari_CI_example}
    \end{subfigure}%
    \caption{Online policy evaluation with TD learning in the Atari Pong RL environment: Figure \ref{fig:atari_pong_example} depicts an example of the Atari Pong game. This is the raw input to the learner, in the form of pixels and RGB color information. Figure \ref{fig:atari_CI_example} shows an example CI generated by the online bootstrap method for the value estimate of the initial state. Both the Quantile (Q) and Standard Error (SE) CIs are included.}
    \label{fig:atari1}
\end{figure}

In this experiment, we consider the problem of deep Q-learning \citep{mnih2015human} in the Atari Pong environment \citep{openai}. Here the agent controls a paddle in a game of pong, competing against a system paddle, as depicted in Figure \ref{fig:atari_pong_example}. At the end of each round, a score of 1 is awarded when a paddle hits the ball past the other paddle. The game is won by the paddle that first reaches a score of 21. 

In terms of the RL environment, a reward of 1 is awarded (deducted) when the agent wins (loses) a round. An episode ends when the game is won or lost. The total reward for the episode is the net score accumulated by the agent. Rewards are scaled by 0.1 before being used as inputs to the RL agent. The state inputs to the agent are the raw pixels (RGB images) generated by the game engine. So each state is an array of shape $(210,160,3)$. There are 6 discrete actions available to the agent at each step.

Our goal here is to learn a linear function approximation of the value function associated with a given policy. However, in this case, since the states are represented as 3-way tensors, we cannot use these states as raw features. Instead, we use a neural network to transform the states into feature vectors. We first train the agent using a Deep Q Network (DQN) with the same configuration as in \cite{mnih2013atari}, with three convolutional layers, and two fully connected hidden layers. The input is the pre-processed state tensor, while the outputs are the Q-function estimates for each state-action pair associated with that state. 

In order to apply our algorithm to this case, we drop the output layer of the pre-trained DQN, so that the output of the resulting network is a 512-dimensional vector. These are the features we use for linear TD learning. The linear function approximation parameter $\theta$ is then a vector in $\R^{512}$. Aside from this feature transformation step, the application of the algorithm is the same as before. Figure \ref{fig:atari_CI_example} shows an example of the confidence intervals generated for the initial state using our algorithm over 100 episodes, depicting both quantile (Q) and standard error (SE) estimators for the confidence intervals.  

\begin{figure}[h]
    \centering
    \begin{subfigure}{0.485\textwidth}
        \includegraphics[width=\hsize]{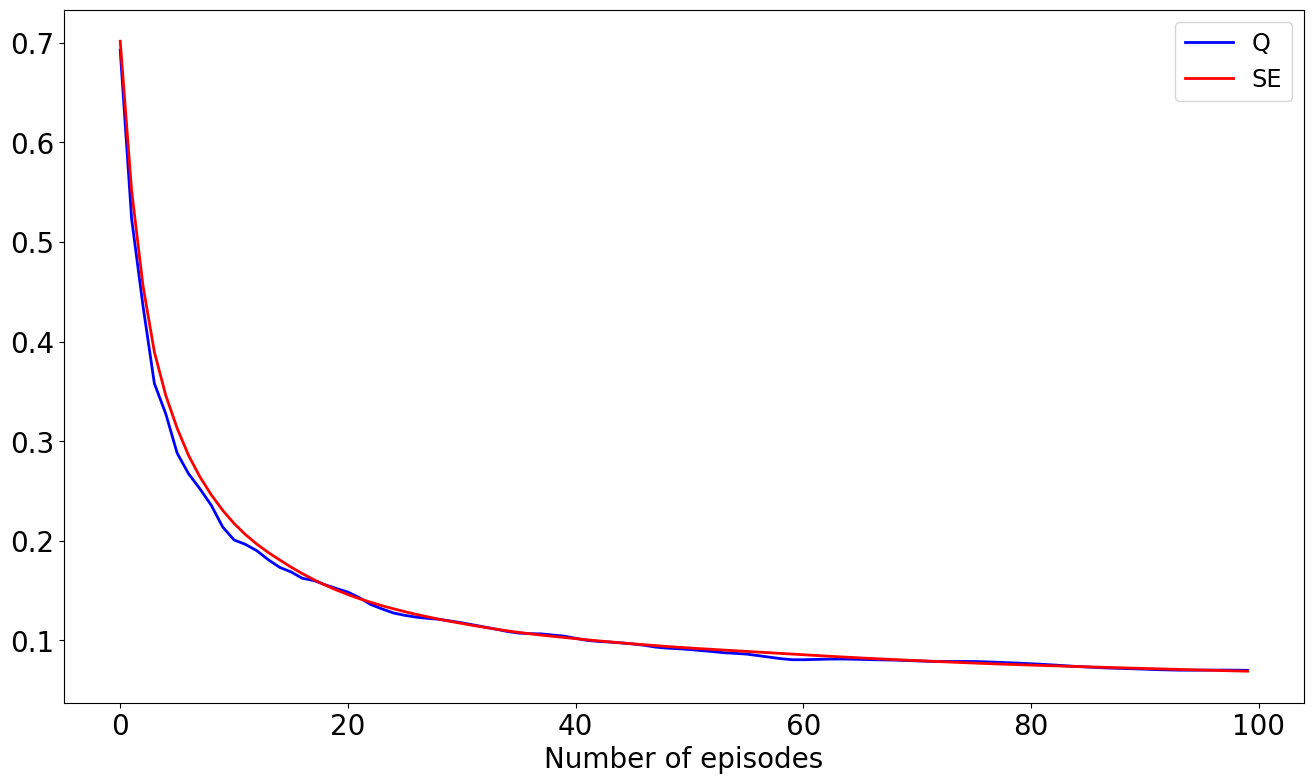}
        \caption{CI Widths}
        \label{fig:atari_CI_widths}
    \end{subfigure}%
    \hspace{0.02\textwidth}%
    \begin{subfigure}{0.485\textwidth}
        \includegraphics[width=\hsize]{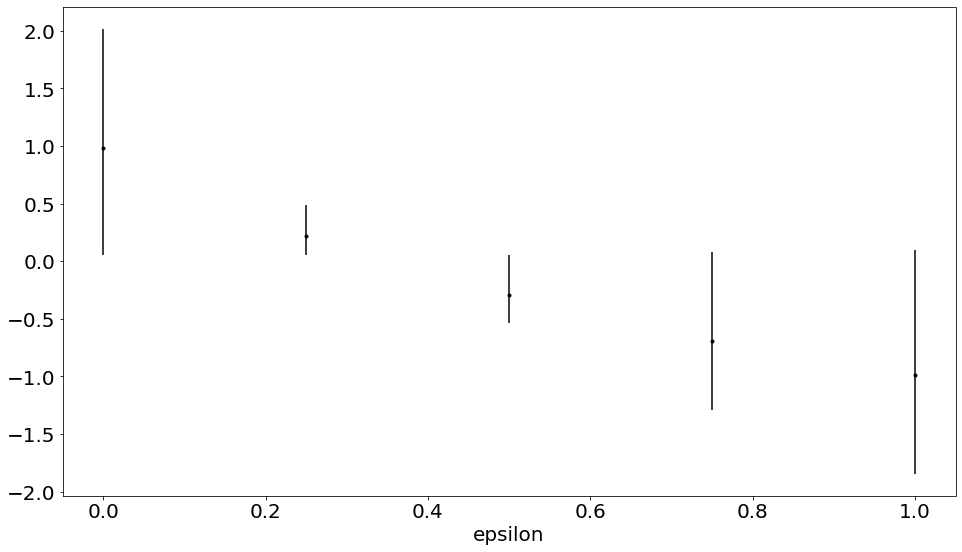}
        \caption{95\% CI Error Bars}
        \label{fig:atari_CI_bars}
    \end{subfigure}%
    \caption{Online policy evaluation with TD learning in the Atari Pong RL environment: Figure \ref{fig:atari_CI_widths} shows the widths of the CIs generated by the Quantile and Standard Error estimators. Figure \ref{fig:atari_CI_bars} shows 95\% CI error bars comparing the value estimates for 5 $\epsilon$-greedy policies derived from the optimal policy ($\epsilon = 0$).}
    \label{fig:atari2}
\end{figure}

Finally, Figure \ref{fig:atari_CI_widths} shows the widths of the confidence intervals computed using these estimators. Figure \ref{fig:atari_CI_bars} shows the error bars of the value estimates for the initial state for a number of $\epsilon$-greedy variants of the policy learnt using deep Q-learning. This figure captures the fact that the mean value estimates decrease gradually as a function of $\epsilon$, with the optimal policy ($\epsilon = 0$) and the random policy ($\epsilon = 1$) having the highest and lowest mean value estimates, respectively. It also shows that the policies with some amount of randomness ($\epsilon > 0$) have lower variance than the optimal policy, which reflects the fact that optimal policies are generally more prone to over-fitting, resulting in higher variance. In all these cases, the same pre-trained network was applied to compute the transformed features, which were then used as inputs to the linear TD algorithm.

\color{black}
\subsection{Off-Policy Value Inference For Infinite-Horizon MDP}
In our next experiment, we consider a simulated infinite-horizon MDP setting for off-policy evaluation. Here, we construct an environment with a state space and action space of size 50 and 5, respectively. The transition probability kernel of the MDP, the state features, and the policy we evaluate are randomly generated. We construct the true value function corresponding to the generated policy as the inner product of the state features and a randomly generated true parameter. The expected rewards at each state under the given policy are then computed using the Bellman equation (\ref{eq:bellman}). Observations are sequentially drawn according to the MDP under the generated policy, and we use this data to estimate the value of an $\epsilon$-greedy version of the generated policy, with $\epsilon = 0.2$.  

Within this framework, we implement three off-policy evaluation methods - the online bootstrap, the offline (vanilla) bootstrap, and the SAVE estimator \citep{shi2021statistical}. We compare the methods in terms of their CI widths and empirical coverage probabilities (computed over 200 runs) for the value estimate of one of the states. Our online bootstrap method is applied in conjunction with GTD learning \citep{sutton09} to obtain the point estimates. This an online method that updates the parameters sequentially. For the offline bootstrap method, to ensure a fair comparison, we use the same GTD estimator to obtain the point estimate, and use the vanilla bootstrap to compute the confidence intervals. A similar offline bootstrap method has been used by \cite{hanna17} and \cite{hao2021bootstrapping} for constructing their confidence intervals. The SAVE method is used here in conjunction with LSTD-Q \citep{lagoudakis2003}, i.e., without the sieve basis functions. Note that LSTD-Q is a batch learning method, and therefore has different properties to GTD learning. 


\begin{figure}[h]
    \centering
    \begin{subfigure}{0.485\textwidth}
        \includegraphics[width=\hsize]{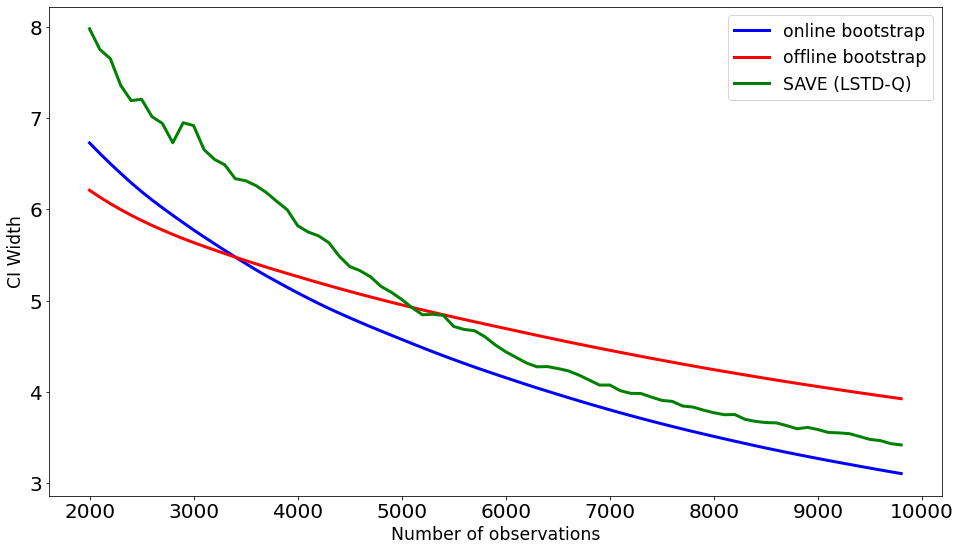}
        \caption{CI Widths}
        \label{fig:offpolicy_compare_widths}
    \end{subfigure}%
    \hspace{0.02\textwidth}%
    \begin{subfigure}{0.485\textwidth}
        \includegraphics[width=\hsize]{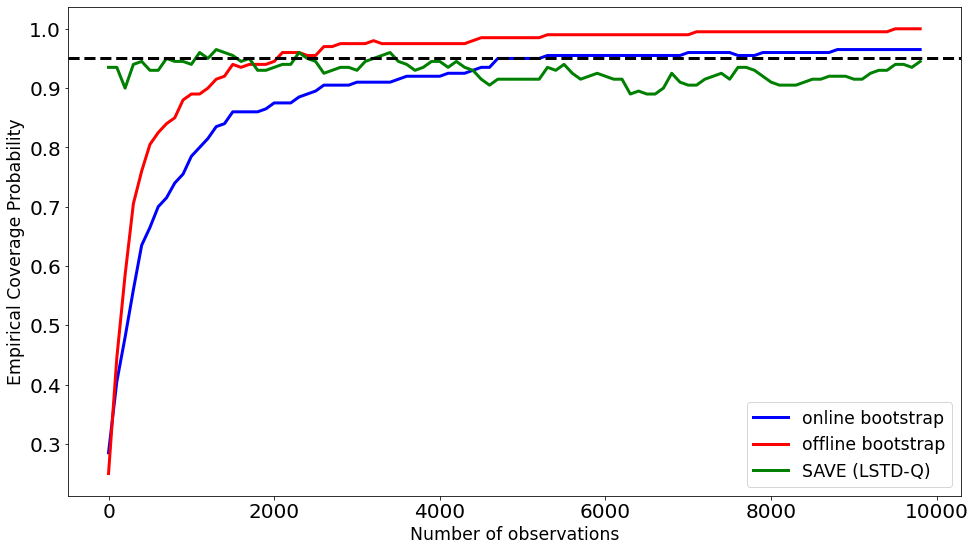}
        \caption{Empirical Coverage Probabilities}
        \label{fig:offpolicy_compare_coverage}
    \end{subfigure}%
    \caption{Comparison of online bootstrap method to offline (vanilla) bootstrap and SAVE (LSTD-Q) methods in the infinite-horizon simulated MDP setting.}
    \label{fig:offpolicy_compare}
\end{figure}

Figure \ref{fig:offpolicy_compare} shows the results of our comparisons. Figure \ref{fig:offpolicy_compare_widths} compares the widths of the confidence intervals, while Figure \ref{fig:offpolicy_compare_coverage} compares the empirical coverage probabilities over 200 simulated runs. The results show that the three methods perform roughly on par, although our online bootstrap method is much faster than the offline bootstrap, and, unlike the other two methods, does not need to store the data. Since the SAVE method uses batch LSTD-Q for its point estimates, it is expected to reach the desired 0.95 coverage probability in fewer observations than the other two online GTD-based methods.

\color{black}
\subsection{Off-Policy Value Inference For A Healthcare Application}
Finally, we consider a real-world example from the healthcare domain, namely, the problem of treating sepsis in the intensive care unit. For this experiment, we use the septic management simulator by \cite{oberst19}. It simulates the patient's vital signs, such as the heart rate, blood pressure, oxygen concentration, and glucose levels. There are three treatment actions (antibiotics, vasopressors, and mechanical ventilation) which the RL agent chooses from at each time step. The reward is +1 if the patient is discharged and -1 if the patient reaches a critical state. These are both absorbing states. The reward is 0 for transitions to non-absorbing states.

As in the previous experiment, we train a near-optimal policy using Q-learning. We then generate a dataset using an $\epsilon$-greedy Q-policy, with $\epsilon = 0.05$. For the target policy, we use the optimal policy learned using Q-learning. To estimate the value function of the target policy in the off-policy setting, we use GTD learning with the Mean-Squared Projected Bellman Error (MSPBE) objective function \citep{sutton2009fast}. 

\begin{figure}[h]
    \centering
    \begin{subfigure}{0.485\textwidth}
        \includegraphics[width=\hsize]{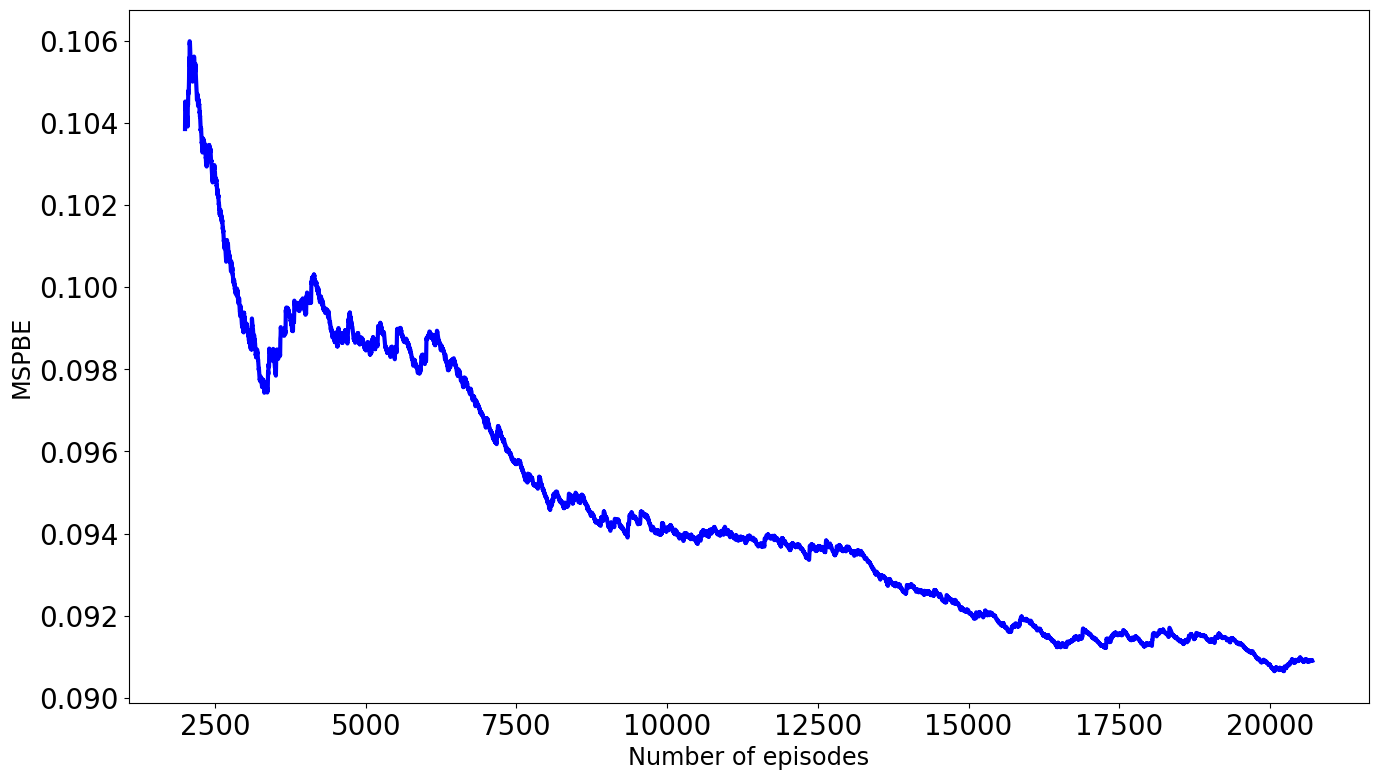}
        \caption{Estimation error}
        \label{fig:septic_sim_MSPBE}
    \end{subfigure}%
    \hspace{0.02\textwidth}%
    \begin{subfigure}{0.485\textwidth}
        \includegraphics[width=\hsize]{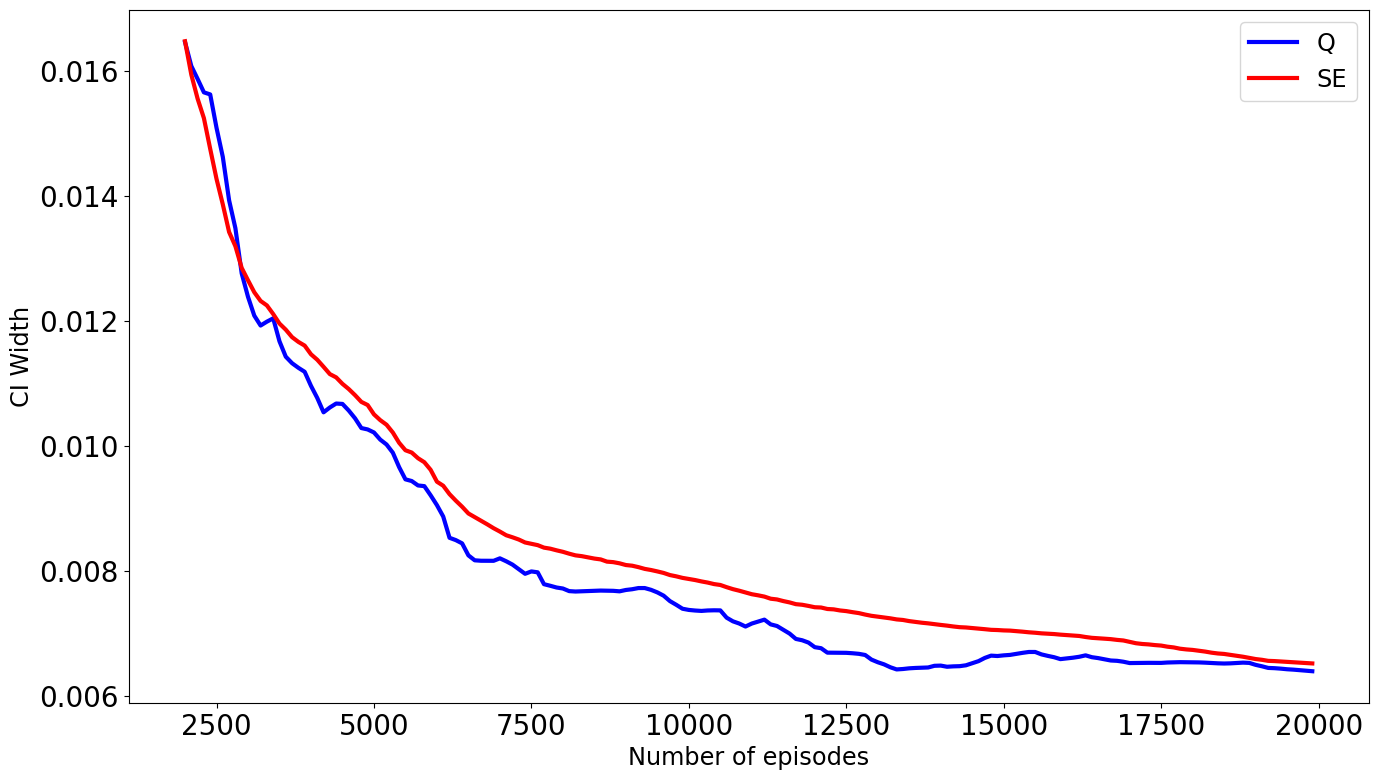}
        \caption{CI widths}
        \label{fig:septic_sim_CI_width}
    \end{subfigure}%
    \caption{Offline policy evaluation with GTD learning in the Septic simulation environment: Figure \ref{fig:septic_sim_MSPBE} shows the estimation error of the GTD learning algorithm, measured in terms of the Mean-Squared Projected Bellman Error (MSPBE). Figure \ref{fig:septic_sim_CI_width} shows the width of the confidence intervals computed using the Quantile (Q) and Standard Error (SE) estimators. }
    \label{fig:septic_sim}
\end{figure}

Since we do not have the true transition matrix, it is not possible to estimate coverage frequencies for our algorithm using the true value function. Figure \ref{fig:septic_sim_MSPBE} shows the empirical MSPBE, which is a proxy for the estimation error for the of the point estimates. Figure \ref{fig:septic_sim_CI_width} shows the width of the confidence intervals computed using the quantile and standard error estimators. The estimation error decreases gradually over 20,000 episodes, and the CI widths for both estimators decrease correspondingly, reflecting the decreasing uncertainty in the estimates as more data becomes available.

\color{black}
\section{Discussion and Future Work}
\label{sec:discussion}

In this paper, we present a  fully online bootstrap algorithm for statistical inference of policy evaluation in reinforcement learning. We establish its distributional consistency in terms of the underlying algorithm, linear stochastic approximation under Markov noise. Our experimental results suggest that the online bootstrap method is efficient and effective across a range of tasks, from linear SGD with Markov noise to off-policy value estimation with GTD learning. Next, we discuss a few additional topics and interesting future directions. 

\subsection{Non-Stationary Behavior Policy}
Our online inference method is built upon linear stochastic approximation, and focuses on two applications in RL: (1) on-policy evaluation with standard TD learning, where the target and behavior policies are the same, and (2) off-policy evaluation with GTD learning, where the target and behavior policies are different. An interesting additional application to consider would be the case where the behavior policy changes over time, as this would enable the estimation of the value of a target policy from observations generated under a non-stationary behavior policy. To our knowledge, all existing stochastic approximation-based off-policy evaluation algorithms (GTD learning: \cite{sutton09}, Emphatic TD: \cite{sutton2016emphatic_td}, and related algorithms) require the behavior policy to be stationary in order to ensure theoretically valid point estimates. That said, if these point methods were proven to be convergent under a non-stationary behavior policy, our online bootstrap-based inference method should easily be adaptable to that setting, as it only requires knowledge of the importance sampling ratio at each time step.

\subsection{Semi-parametric Efficiency}

The main purpose of this paper is to propose a provable online bootstrap inference method for the existing point estimators - TD learning and GTD learning, which are both special cases of linear stochastic approximation under Markov noise. In the context of i.i.d. observations, the asymptotic variance of the averaged iterate under this scheme is known to achieve the Cramer-Rao lower bound \citep{polyak92, moulines2011}. 

There are a few recent works studying the semi-parametric efficiency of reinforcement learning algorithms. For example, \cite{ueno2011generalized} proposed a generalized form of TD learning and studied its semi-parametric efficiency. Using this framework, they derived an optimal estimating function with the minimal asymptotic variance. In addition, a recent work by \cite{kallus2021} established a lower bound for the asymptotic mean-squared error (MSE) of the Q-function estimate under a general function approximation scheme. They provide an explicit form for the approximating function that achieves the MSE lower bound. Their results require an assumption of transition sampling, i.e., state-action-reward-state tuples are drawn independently from the generative model, rather than sequentially according to the trajectory of the MDP. 

To our knowledge, there are currently no results for the efficiency of the linear stochastic approximation iterate under Markov noise. Since our focus is on the bootstrap method for uncertainty quantification, rather than the algorithm for point estimation, we leave the question of efficiency for linear stochastic approximation under Markov noise as interesting future work. 

\subsection{Extension to Non-Linear Function Approximation}
In this work we mainly focus on the linear function approximation setting, as this is the fundamental function approximation scheme for policy evaluation algorithms in RL, serving as the basis for more elaborate approximation schemes. Here we briefly discuss two practical ways in which we can use the online bootstrap approach within the non-linear function approximation setting.

The first approach is to transform the feature basis to another space using nonlinear basis expansion, and then use bootstrapping in conjunction with linear TD learning on the transformed basis. The basis expansion can be done via sieve basis functions \citep{shi2021statistical} or neural networks to learn the representation of the feature space. In our Atari Deep Q-learning experiment of Section 5.2, the raw state features are high-dimensional 3-way tensors. Here, we applied linear TD learning on the features obtained from the last layer of the pre-trained DQN to handle the non-linear function approximation. This is similar to the approach used by \cite{chung2019}, who proposed a two-timescale network architecture that enables linear methods to learn values at the top layer, with a non-linear representation learned at a slower timescale at the bottom layers.

The second approach is to directly apply our method to a non-linear function approximation scheme. The general form of the stochastic approximation is
\begin{align}
    \theta_{t+1} = \theta_t + \alpha_{t+1} H(\theta_t, X_{t+1}), \label{eq:nlsa1}
\end{align}
where $H(\theta,X)$ is a noisy observation of the mean field $h(\theta) = \mathbb E[H(\theta, X)]$. Here, the goal is to estimate the root $\theta_*$ of the non-linear equation $h(\theta) = 0$. Under some assumptions on the non-linear function $h$ and the iterates $\{\theta_t \}$, \cite{andrieu05} proved the consistency of the $\{\theta_t\}$, while \cite{liang2010} established a central limit theorem for the averaged iterate $\thetabar_t$. In practice, we can apply our online bootstrap procedure in Algorithm 1 to produce bootstrap estimates for the iterates $\thetabar_t$ under the non-linear stochastic approximation update (\ref{eq:nlsa1}). This allows us to perform online inference for non-linear SGD under Markov noise, which extends the online inference results for non-linear SGD under i.i.d. noise \citep{fang19, chen20sgd}. Since the focus of this work is on linear function approximation, we leave a rigorous investigation of this non-linear setting to future work.

\subsection{Approximation Error and Model Misspecification}
This paper focuses on the linear function approximation of the value function. When the linear model assumption is violated, it is important to consider the model mis-specification issue in the analysis of RL algorithms with function approximation. Here we briefly discuss the implications of approximation error within the context of TD learning with linear function approximation. Our discussion starts from the least-false parameter in the linear space and then connects it with the point estimator from TD learning. 

\textbf{Least-false parameter in the linear space:} We follow the notation defined in Section 2 of the paper. Denote $\Scal$ as the state space, $\Phi$ as the feature matrix. Let $\Pi$ denote the projection operator onto the space spanned by the linear basis functions. Then, for a given policy $\pi$ with a stationary distribution $\mu$, we have
\begin{align*}
    \Pi V^\pi = \argmin_{\overline{V} \in \{\Phi \theta | \theta \in \R^d \}}\left\| V^\pi - \overline{V} \right\|_D,
\end{align*}
where $D$ denotes the diagonal matrix with elements corresponding to the entries of the stationary distribution $\mu$, and $\| v \| = \sqrt{v\Tra D v}$ denotes the norm under the stationary distribution. In other words, the projected value function $\Pi V^\pi$ is the best approximation to the true value function $V^\pi$ within the subspace spanned by the linear basis functions.

\textbf{Point estimator from TD learning:} \cite{tsitsiklis1997} showed that the limiting point $\theta_*$ of the linear TD update is the unique solution to the projected Bellman equation $V_\theta = \Pi T^{\pi} V_\theta$, where $V_\theta = \Phi \theta$ is the value estimate corresponding to the parameter $\theta$, while $T^\pi$ denotes the Bellman operator under the policy $\pi$. In other words, the limiting point $\theta_*$ of the TD estimator can be seen as the global minimizer of the Mean-Squared Projected Bellman Error (MSPBE), i.e.,
\begin{align*}
    \theta_* = \argmin_{\theta \in \R_d} \left\| V_\theta - \Pi T^\pi V_\theta \right\|_D^2.
\end{align*}
Our online bootstrap method provides a way to perform inference on the minimizer of the MSPBE, which is a quantity of interest in its own right \citep{sutton2018reinforcement}. 

\textbf{Connection:} Although the limiting point of the linear TD estimator is not the least-false parameter in the linear space, these two have a nice connection. \cite{tsitsiklis1997} showed that the value function $V_{\theta_*}$ corresponding to the limiting point $\theta_*$ satisfies
\begin{align*}
    \left\| V_{\theta_*} - V^\pi \right\|_D \leq \frac{1}{\sqrt{1 - \gamma^2}} \left\| \Pi V^\pi - V^\pi \right\|_D.
\end{align*}
Thus, the approximation error for the limiting point of the TD value estimator is bounded by a constant times the approximation error for the projected value function $\Pi V^\pi$, which represents the best possible approximation in the span of $\Phi$. So, while the confidence intervals generated by the online bootstrap method provide a coverage only for the value function $V_{\theta_*}$ corresponding to the minimizer of the MSPBE, this value function itself comes with a competitive guarantee.

\color{black}

\section*{Acknowledgment}
The authors thank the editor Professor Marina Vannucci, the associate editor and three anonymous reviewers for their valuable comments and suggestions which led to a much improved paper. Zhaoran Wang acknowledges National Science Foundation (Awards 2048075, 2008827, 2015568, 1934931), Simons Institute (Theory of Reinforcement Learning), Amazon, J.P. Morgan, and Two Sigma for their supports. Will Wei Sun's research was partially supported by ONR grant N00014-18-1-2759.  Guang Cheng acknowledges support from the National Science Foundation (NSF -- SCALE MoDL (2134209)). Any opinions, findings, and conclusions or recommendations expressed in this material are those of the authors and do not necessarily reflect the views of National Science Foundation and Office of Naval Research. The authors report there are no competing interests to declare.

\baselineskip=13pt
\bibliographystyle{chicago}
\bibliography{refs}
\newpage
\baselineskip=24pt
\setcounter{page}{1}
\setcounter{equation}{0}
\setcounter{section}{0}
\renewcommand{\thesection}{S\arabic{section}}
\renewcommand{\theequation}{\thesection.\arabic{equation}}

\begin{center}
{\Large\bf SUPPLEMENTARY MATERIAL} \\
\medskip
{\Large\bf Online Bootstrap Inference For Policy Evaluation in Reinforcement Learning} \\
\end{center}

In this online supplementary material, we provide detailed proofs for the lemmas and main theorems in Sections \ref{sec:theory_lsa} and \ref{sec:theory_plsa}, as well as additional experiments.

\section{Proofs for Section \ref{sec:theory_lsa}}
\label{apdx1}
\color{black}
The following lemma is a restatement of Theorem 2 from \cite{chong1999sa}. In the following, we say that a sequence $\epsilon_t$ is \textit{small} with respect to another sequence $\alpha_t$ if there exist sequences $\{ e_t\}$ and $\{ r_t\}$ such that $\epsilon_t = e_t + r_t$ for all $t$, $r_t \to 0$, and $\sum_{k=1}^t \alpha_t \| e_t \|_2 $ converges. Also, we say that a scalar sequence $\{ a_t \}$ has \textit{bounded variation} if $\sum_{t = 1}^\infty | a_{t+1} - a_t | < \infty$.
We refer to the cited article for further details on these conditions.

\begin{lemma} Consider the linear stochastic approximation update
\label{lem:cwk99}
\begin{align*}
    \theta_{t+1} = \theta_t + \alpha_{t+1} (\Atilde(X_{t+1}) \theta_t - \btilde(X_{t+1})).
\end{align*}
Assume the following conditions hold:
\begin{enumerate}[({B}1)]
    \item The step size sequence $\{\alpha_t\}$ satisfies $\alpha_t > 0$, $\alpha_t \to 0$, and $\sum_{t=1}^\infty \alpha_t = \infty$.
    \item $\Abar$ is a bounded Hurwitz matrix.
    \item $\{\Atilde(X_t) - \Abar\}$ is small with respect to $\alpha_t$.
    \item Let $\{\rho_t\}$ be a positive real sequence converging monotonically to 0, such that
    \begin{enumerate}[(i)]
        \item $\{\rho_t^{-1}(\btilde(X_t) - \bbar)\}$ is small with respect to $\alpha_t$.
        \item $(\rho_t - \rho_{t+1})/(\alpha_t \rho_t) \to c < \infty$,
        \item The sequences $\{ \rho_{t+1}/\rho_t \}$ and $\{\rho_t / \rho_{t+1}\}$ have bounded variation.
    \end{enumerate}
\end{enumerate}
Then $\theta_t - \theta_* = o(\rho_t)$.
\end{lemma}

\subsubsection*{Proof of Proposition \ref{thm:lsa_as}}
We verify that the conditions of Lemma \ref{lem:cwk99} hold under our assumptions \ref{asu1}, \ref{asu2}, \ref{asu3}. 

Firstly, it is easy to see that (B1) holds under \ref{asu3}, i.e., with a step size $\alpha_t = \alpha_0 t^{-\eta}$, $\eta \in (1/2,1)$. Similarly, (B2) follows directly from assumption \ref{asu2}.

For functions $f$ defined over the state space $\Xcal$, we define the $t$-step transition operator $\Pcal^t f(x) = \int_{y \in \Xcal} f(y) \Pcal^t(x,dy)$, where $\Pcal^t(x,y)$ denotes the $t$-step transition probability from state $x$ to $y$. When $t = 1$, we write $\Pcal^1 f(x) = \Pcal f(x)$.

Next, define $\Ahat: \Xcal \to \R^{d \times d}$ and $\bhat: \Xcal \to \R^d$ to be the solutions to the Poisson equations
\begin{align*}
    \Atilde(x) - \Abar(x) &= \Ahat(x) - \Pcal \Ahat(x), \\
    \btilde(x) - \bbar(x) &= \bhat(x) - \Pcal \bhat(x),
\end{align*}
for $x \in \Xcal$. The existence of $\Ahat$ and $\bhat$ is guaranteed under \ref{asu1}. Furthermore, under \ref{asu2}, there exist constants $\Ahat_{\max}, \bhat_{\max > 0}$ such that $\|\Ahat\|_F \leq \Ahat_{\max}$ and $\|\bhat\|_2 \leq \bhat_{\max}$ \citep{delyon2000stochastic}.

Then we can write $\Atilde(X_t) - \Abar = e_t + r_t$, where $e_t = \Ahat(X_t) - \Pcal \Ahat(X_{t+1})$, and $r_t = \Pcal \Ahat(X_{t+1}) - \Pcal \Ahat(X_{t})$. To verify condition (B3), it suffices to show that $\sum_{t=1}^\infty \alpha_t \|e_t\| < \infty$, and $\|r_t\| \to 0$, as $t \to \infty$ \citep{chong1999sa}. 

Let $\Fcal_t = \sigma(\{X_t\})$ denote the natural filtration with respect to the Markov chain $X_t$. Then $\E[e_t | \Fcal_t] = 0$, and $e_t$ is a martingale difference sequence with respect to $\Fcal_t$. Furthermore, $e_t$ is a.s. uniformly bounded, since $\|e_t\|_F \leq 2 \Ahat_{\max}$, by construction. So $\sum_{t=1}^\infty \alpha_t^2 \E[\|e_t\|^2 | \Fcal_t] < \infty$. Then, by Theorem 29 of \cite{delyon2000stochastic}, $\sum_{t=1}^\infty \alpha_t \| e_t \|$ converges.

Also, since $\Pcal(X_t, \cdot) \to \mu$ as $t \to \infty$, it follows that $\|r_t\| \to 0$, as $t \to \infty$. So $\{\Atilde(X_t) - \Abar\}$ is small with respect to $\alpha_t$, and (B3) holds.

Next, set $\rho_t = t^{-\gamma}$, with $\gamma \in (0, \eta-1/2)$. Conditions (B4)(ii) and (B4)(iii) hold under this definition, with $c = 0$ in (B4)(ii) \citep{chong1999sa}. It remains to verify (B4)(i).

Define $\btilde(X_t) - \bbar = e_t + r_t$, where $e_t = \bhat(X_t) - \Pcal \bhat(X_{t+1})$, and $r_t = \Pcal \bhat(X_{t+1}) - \Pcal \bhat (X_t)$. It suffices to show that $\sum_{t=1}^\infty \alpha_t \rho_t^{-1} \|e_t\| < \infty$, and that $\rho_t^{-1} \|r_t\| \to 0$, as $t \to \infty$.

By the same argument used above for $\{\Atilde - \Abar\}$, $e_t$ is an a.s. uniformly bounded martingale difference sequence, with $\|e_t\|_F \leq 2 \bhat_{\max}$ for all $t$. Then $\sum_{t=1}^\infty \alpha_t^2 \rho_t^{-2} \E[\|e_t\|_2 | \Fcal_t] < \infty$, since $\eta - \gamma > 1/2$. So, by Theorem 29 of \cite{delyon2000stochastic}, $\sum_{t=1}^\infty \alpha_t \rho_t \|e_t\|$ converges.

Next, we have
\begin{align*}
    \| \Pcal \bhat(X_{t+1}) - \Pcal \bhat (X_t) \| & = \left\| \int_{y \in \Xcal} \bhat(y) (\Pcal(X_{t+1}, dy) - \Pcal(X_{t}, dy) )\right\| \\
    & \leq \int_{y \in \Xcal} \| \bhat(y) \| \| \Pcal(X_{t+1}, dy) - \Pcal(X_{t}, dy) \| \\
    & \leq \bhat_{\max} \int_{y \in \Xcal} \| \Pcal(X_{t+1}, dy) - \Pcal(X_{t}, dy) \|.
\end{align*}
Consider the integrand in the above expression. For any bounded initial distribution $\nu_0$, i.e., with $\sup_{x \in \Xcal} \|\nu_0(x)\| \leq \nu_{\max}$, for some constant $\nu_{\max} < \infty$, we have
\begin{align*}
    \| \Pcal(X_{t+1}, dy) - \Pcal(X_{t}, dy) \| & \leq \sup_{x_1, x_2 \in \Xcal} \| \nu_0 \Pcal^{t+1}(x_1, dy) - \nu_0 \Pcal^{t}(x_2, dy) \| \\
    & \leq \nu_{\max} \sup_{x_1, x_2 \in \Xcal} \| \Pcal^{t+1}(x_1, dy) - \Pcal^{t}(x_2, dy) \| \\
    & = \nu_{\max} \sup_{x_1, x_2 \in \Xcal} \| (\Pcal^{t+1}(x_1, dy) - \pi(dy)) - (\Pcal^{t}(x_2, dy) - \pi(dy)) \| \\
    & \leq \nu_{\max} \sup_{x_1, x_2 \in \Xcal} \left( \| \Pcal^{t+1}(x_1, dy) - \pi(dy)\| + \| \Pcal^{t}(x_2, dy) - \pi(dy)\| \right) \\
    & \leq \nu_{\max} \left( M \kappa^{t+1} + M \kappa^{t} \right) \\
    & \leq 2 \nu_{\max} M \kappa^t,
\end{align*}
where the penultimate inequality holds by (\ref{eq:mixing}). It follows that $\| \Pcal \bhat(X_{t+1}) - \Pcal \bhat (X_t) \| \leq 2 \bhat_{\max} \nu_{\max} M \kappa^t$, and so $\rho_t^{-1} \| r_t \| \leq 2 \bhat_{\max} \nu_{\max} M t^{\gamma} \kappa^t \to 0$, as $t \to \infty$. \qed

\subsubsection*{Proof of Proposition \ref{thm:lsa_clt}}
First, we list the conditions required for our central limit theorem, Proposition \ref{thm:lsa_clt}, to hold. The assumptions listed below are from \cite{liang2010}, who proved a central limit theorem for the varying truncation stochastic approximation MCMC algorithm. This is a general form of algorithm (\ref{eq:lsa1}), and is designed to solve the equation
\begin{align*}
    h(\theta) = \int_{\Xcal}H(\theta,x) f_\theta(x) dx = 0,
\end{align*}
where $\theta \in \Theta \subset \R^{d_\theta}$ is a parameter vector and $f_\theta(x), x \in \Xcal \subset \R^{d_x}$ is a density function depending on $\theta$. The function $h(\theta)$ is called the mean field function, and $H(\theta,x)$ is a noisy observation of $h(\theta)$. 

The stochastic approximation algorithm is designed to iteratively estimate $\theta$ from a sequence of noisy observations that depend on the current estimate of $\theta$ (hence forming a controlled Markov chain). The main update step for this algorithm is given by
\begin{align}
    \theta_{t+1} & = \theta_t + \alpha_{t+1} H(\theta_t, X_{t+1}) \nonumber \\
    & = \theta_t + \alpha_{t+1} h(\theta_t) + \alpha_{t+1} \epsilon_{t+1}, \label{eq:liang_sa}
\end{align}
where $h(\theta) = \int H(\theta,x) f_\theta(x) dx$, $f_\theta$ being the invariant distribution of the controlled Markov transition kernel $\Pcal_\theta$, and $\epsilon_{t+1} = H(\theta_t, X_{t+1}) - h(\theta_t)$ is the residual noise term.

In order to ensure the convergence of the iterates in (\ref{eq:liang_sa}), \cite{liang2010} imposes a varying truncation scheme, whereby the iterates $\theta_t$ are constrained within an increasing sequence of compact sets $\{ \Kcal_s \}_{s \geq 0}$. Under this scheme, \cite{andrieu05} showed that there exists a time step $t_{\sigma_s} < \infty$ such that $\theta_t \in \Kcal_{\sigma_s}$ for all $t \geq t_{\sigma_s}$, and there are no further truncations beyond time step $t_{\sigma_s}$. The central limit theorem applies to the averaged iterate $\thetabar_t := \frac{1}{t - t_{\sigma_s}}\sum_{i = t_{\sigma_s} + 1}^t \theta_i$.  

The following conditions are assumed by \cite{liang2010}:

\begin{enumerate}[({C}1)]
    \item $\Theta$ is an open set, the function $h: \Theta \to \R^d$ is continuous, and there exists a continuously differential function $v: \Theta \to [0,\infty)$ such that
    \begin{enumerate}[(i)]
        \item There exists $M_0 > 0$ such that
        \begin{align*}
            \Lcal = \{\theta \in \Theta, \langle \nabla v(\theta), h(\theta) \rangle = 0 \} \subset \{\theta \in \Theta, v(\theta) < M_0 \}.
        \end{align*}
        \item There exists $M_1 \in (M_0, \infty)$ such that $\mathcal{V}_{M_1}$ is a compact set, where $\mathcal{V}_M = \{\theta \in \Theta, v(\theta) \leq M \}$.
        \item For any $\theta \in \Theta \backslash \Lcal$, $\langle \nabla v(\theta), h(\theta) \rangle < 0$.
        \item The closure of $v(\Lcal)$ has an empty interior.
    \end{enumerate}
    
    \item The mean field $h(\theta)$ is measurable and locally bounded. There exists a Hurwitz matrix $F$, $\gamma > 0, \rho \in (0,1]$, and a constant $c$ such that, for any $\theta_* \in \Lcal$,
    \begin{align*}
        \| h(\theta) -F(\theta - \theta_*)\| \leq c\| \theta - \theta^*\|^{1+\rho} \quad \forall \theta \in \{ \theta: \| \theta - \theta_*\| \leq \gamma\},
    \end{align*}
    where $\Lcal$ is defined in (B1)(i).
    
    \item For any $\theta \in \Theta$, the transition kernel $\Pcal_\theta$ is irreducible and aperiodic. In addition, there exists a function $V: \Xcal \to [1,\infty)$, and a constant $\alpha \geq 2$ such that for any compact set $\Kcal \subset \Theta$:
    \begin{enumerate}[(i)]
        \item There exists a set $\mathbf{C} \subset \Xcal$, and integer $l$, constants $0 < \lambda < 1, b, \zeta, \delta > 0$ and a probability measure $\nu$ such that
        \begin{align*}
            & \sup_{\theta \in \Kcal} \Pcal_\theta^l V^\alpha(x) \leq \lambda V^\alpha(x) + b I(x \in \mathbf{C}) \quad \forall x \in \Xcal, \\
            & \sup_{\theta \in \Kcal} \Pcal_\theta V^\alpha(x) \leq \zeta V^\alpha(x) \quad \forall x \in \Xcal, \\
            & \inf_{\theta \in \Kcal}\Pcal_\theta^l(x,A) \geq \delta \nu(A) \quad \forall x \in \mathbf{C}, \forall A \in \Bcal_\Xcal.
        \end{align*}
        \item There exists a constant $c > 0$ such that, for all $x \in \Xcal$,
        \begin{align*}
            & \sup_{\theta \in \Kcal}\| H(\theta,x)\|_V \leq c, \\
            & \sup_{\theta, \theta' \in \Kcal}\|H(\theta,x) - H(\theta',x)\|_V \leq c\|\theta - \theta'\|.
        \end{align*}
        \item There exists a constant $c > 0$ such that, for all $\theta, \theta' \in \Kcal$,
        \begin{align*}
            & \|\Pcal_\theta g - \Pcal_{\theta'}\|_V \leq c\|g\|_V \|\theta - \theta'\| \quad \forall g \in \Lcal_V, \\
            & \| \Pcal_\theta g - \Pcal_{\theta'} g \|_{V^\alpha} \leq c \|g \|_{V^\alpha} \|\theta - \theta'\| \quad \forall g \in \Lcal_{V^\alpha}.
        \end{align*}
    \end{enumerate}
    
    \item The step sizes $\{\alpha_t\}$ are non-increasing, positive sequences that satisfy the conditions
    \begin{gather*}
        \sum_{t=1}^\infty \alpha_t = \infty, \quad \lim_{t \to \infty}(t \alpha_t) = \infty, \quad \frac{\alpha_{t+1}-\alpha_t}{\alpha_t} = o(\alpha_{t+1}), \quad \sum_{t=1}^\infty \frac{\alpha_t^{(1+\tau)/2}}{\sqrt t} < \infty,
    \end{gather*}
 for some $\tau \in (0,1]$ and a constant $\alpha \geq 2$ defined in (B3).
\end{enumerate}

We refer to \cite{liang2010} for further details on these conditions.

We now verify that (C1)-(C4) hold under assumptions \ref{asu1}-\ref{asu3}. We also show that, under our assumptions, the iterates of the update (\ref{eq:lsa1}) are constrained within a compact set $\Kcal \subset \Theta$, thereby avoiding the need for the varying truncation scheme. Then the result directly follows. Using the notation of (\ref{eq:liang_sa}), in our case, we have $H(\theta,x) = \Atilde(x) \theta - \btilde$, with mean field $h(\theta) = \Abar \theta - \bbar$. By \ref{asu2iii}, we have $\E_{X \sim \mu}[H(\theta,X)] = h(\theta)$, for all $\theta \in \Theta$.

(C1) assumes the existence of a global Lyapunov function $v$. We may choose $v(\theta) = \theta\Tra \bbar - \frac{1}{2}\theta\Tra \Abar \theta$. Then $v$ is a global Lyupanov function for the mean field $h$ \citep{andrieu05, liang2010}. $\Lcal$ denotes the set of all valid solutions $\theta^*$ for the equation $h(\theta) = 0$. In our case, since $\Abar$ is Hurwitz by \ref{asu2ii}, there exists a unique solution $\theta^*$ for the linear system $\Abar \theta = \bbar$, and $\Lcal$ is a singleton set. 

For (C2), the measurability and local boundedness of $h$ follows directly from linearity. For the latter part, we may choose $F = A$. Then for any $\theta \in \Theta$, we have $\|h(\theta) - F(\theta - \theta^*)\| \equiv 0$, so (C2) holds.

For (C3), in our case the function $H(\theta,x) = \Atilde(x) \theta - \btilde(x)$ is bounded by \ref{asu2ii}, so we can choose the drift function $V \equiv 1$. Then the first two conditions of (C3)(i) hold trivially. 

The third condition in (C3)(i) is a standard assumption in the Markov Chain Monte Carlo (MCMC) literature, and is referred to as the minorization condition. By Theorem 5.2.2 of \cite{meyn09}, for $\varphi$-irreducible Markov chains, ``small sets'' for which the minorization condition holds exist. By \ref{asu1}, the Markov chain is irreducible, and so, by definition, is $\varphi$-irreducible for some irreducibility measure $\varphi$. Hence the condition holds in our case. 

(C3)(ii) follows directly from \ref{asu2ii}. (C3)(iii) does not apply in our case as we are dealing with a homogeneous Markov chain that does not depend on $\theta_k$ (not have a controlled Markov chain). The conditions of (C4) hold trivially under \ref{asu3}. 

Finally, by Proposition \ref{thm:lsa_as}, we can choose a large enough constant $R_{\theta} > 0$ such that $\|\theta_t - \theta_*\|_2 \leq R_{\theta}$ for all $t \geq 0$. Then $\theta_t \in \Kcal$ for all $t \geq 0$, for some compact set $\Kcal \subset \Theta$.
\qed

\section{Proofs for Section \ref{sec:theory_plsa}}
\label{apdx2}
\subsubsection*{Proof of Proposition \ref{thm:plsa_as}}
To verify the conditions for Lemma \ref{lem:cwk99}, it suffices to check that $\{W_t\Atilde(X_t) - \Abar\}$ and $\{\rho_t^{-1}(W_t \btilde(X_t) - \bbar) \}$ are small with respect to the step sizes $\{\alpha_t\}$. By \ref{asu2ii} and the boundedness of $W_t$, we have $\| W_t \Atilde(X_t) \|_F \leq W_{\max} A_{\max} < \infty$, and $\| W_t \btilde(X_t) \| \leq W_{\max} b_{\max} < \infty$, for all $t \geq 1$. By independence of $W_t$, we have $\E_\mu[W_t \Atilde(X_t) - \Abar] = 0$ and $\E_\mu[W_t \btilde(X_t) - \bbar] = 0$. The rest of the argument is identical to the proof of Proposition \ref{thm:lsa_as}. \qed

\begin{lemma}
\label{lem:a2}
Assume \ref{asu1}-\ref{asu3} hold. Then
\begin{align*}
    \sqrt{t} (\bar{\thetahat}_t - \theta_*) = -\bar{A}^{-1} \frac{1}{\sqrt t}\sum_{i=1}^t \hat{\epsilon}_{i+1} + o_p(1).
\end{align*}
\end{lemma}

\noindent \textit{Proof.} An argument similar to above may be used to verify that conditions (C1)-(C4) also hold for the perturbed SA update (\ref{eq:plsa1}) under assumptions \ref{asu1}-\ref{asu3}. Then the result follows as an intermediate step in the proof of Theorem 2.2 by \cite{liang2010}. \qed

\vspace{0.3cm}
\color{black}
The following lemma is adapted from Lemma 5 of \cite{xu20}.

\begin{lemma}
\label{lem:unif_mixing}
Assume \ref{asu1}-\ref{asu3} hold. Then, for any $i > j$, we have
\begin{align*}
    \left\| \E\left[\Atilde(X_i) | \Fcal_j \right] - \Abar \right\|_F \leq A_{\max} M \kappa^{i-j},
\end{align*}
where $M$ and $\kappa$ refer to the constants from (\ref{eq:mixing}).
\end{lemma}

\noindent\textit{Proof}. By (\ref{eq:mixing}), for any $i > j$, the following holds: 
\begin{align}
    \left\| \Pcal^{i-j}(\cdot | \Fcal_j) - \mu \right\| \leq M \kappa^{i-j}, \label{eq:unif_mixing}
\end{align}
Then we have
\begin{align*}
    \left\| \E\left[ \Atilde(X_i) | \Fcal_j \right] - \Abar \right\|_F & = \left\| \int_{x \in \Xcal} \Atilde(x) \Pcal^{i-j}(dx | \Fcal_j) - \int_{x \in \Xcal} \Atilde(x) \mu(dx) \right\|_F \\
    & \leq \int_{x \in \Xcal} \left\| \Atilde(x) \Pcal^{i-j}(dx | \Fcal_j) - \Atilde(x) \mu(dx)  \right\|_F \\
    & \leq \int_{x \in \Xcal} \left\| \Atilde(x) \right\|_F \left\| \Pcal^{i-j}(dx | \Fcal_j) - \mu(dx) \right\| \\
    & \leq A_{\max} M \kappa^{i-j}.
\end{align*}
The first equality follows from the definition of $\Abar$ in \ref{asu2i}, the second step holds by Jensen's inequality, and the final step follows from \ref{asu2iii} and (\ref{eq:unif_mixing}). \qed

\color{black}
\subsubsection*{Proof of Lemma \ref{lem:plsa2}}
Starting with (\ref{eq:plsa2}), we have
\begin{align}
\label{eq:eps}
    \hat{\epsilon}_{t+1} & = (W_{t+1}\Atilde(X_{t+1}) - \Abar)\thetahat_t - (W_{t+1}\btilde(X_{t+1}) - \bbar) \nonumber \\
    & = W_{t+1}(\Atilde(X_{t+1}) \theta_* - \btilde(X_{t+1})) + (W_{t+1} \Atilde(X_{t+1}) - \Abar)(\thetahat_{t} - \theta_*).
\end{align}
using the fact that $\Abar \theta_* = \bbar$.

By Lemma \ref{lem:a2} and (\ref{eq:eps}), we have
{\small
\begin{align*}
    & \sqrt{t} (\bar{\thetahat}_t - \theta_*) = -\Abar^{-1} \frac{1}{\sqrt t} \sum_{i=1}^t \hat{\epsilon}_{i+1} + o_p(1) \\
    & = -\Abar^{-1}\frac{1}{\sqrt t}\sum_{i=1}^t W_{i+1}(\Atilde(X_{i+1})\theta_* - \btilde(X_{i+1})) - \Abar^{-1}\frac{1}{\sqrt t}\sum_{i=1}^t(W_{i+1} \Atilde(X_{i+1}) - \Abar)(\thetahat_{i} - \theta_*) + o_p(1).
\end{align*}
}

Consider the second term in the above expression. We want to show that this term is $o_p(1)$. It suffices to show that its second moment vanishes as $t \to \infty$. First we expand the second moment and split it into square and cross terms. We have
{
\begin{align*}
    & \quad \E\left[ \left\|\frac{1}{\sqrt t}\sum_{i=1}^t \left(W_{i+1} \Atilde(X_{i+1}) - \Abar \right) \left( \thetahat_{i} - \theta_* \right) \right\|^2_2 \right] \\
    & = \frac{1}{t} \sum_{i=1}^t \sum_{j=1}^t \E\left[ \left\langle \left(W_{i+1} \Atilde(X_{i+1}) - \Abar\right)\left(\thetahat_{i} - \theta_*\right), \left(W_{j+1} \Atilde(X_{j+1}) - \Abar\right)\left(\thetahat_{j} - \theta_*\right) \right\rangle \right] \\
    & = \frac{1}{t}\sum_{i=1}^t \E \left[ \left\| \left(W_{i+1} \Atilde(X_{i+1}) - \Abar\right) \left(\thetahat_{i} - \theta_*\right) \right\|^2_2 \right] \\ & + \frac{1}{t}\sum_{i \neq j} \E\left[\left\langle \left( W_{i+1} \Atilde(X_{i+1}) - \Abar \right) \left( \thetahat_{i} - \theta_* \right), \left(W_{j+1} \Atilde(X_{j+1}) - \Abar \right) \left( \thetahat_{j} - \theta_* \right) \right\rangle \right] \\
    & = I_1 + I_2.
\end{align*}
}

We deal with each term separately. First, we have
\begin{align*}
    I_1 & = \frac{1}{t}\sum_{i=1}^t \E \left[ (\thetahat_{i} - \theta_*)\Tra (W_{i+1} \Atilde(X_{i+1}) - \Abar)\Tra (W_{i+1} \Atilde(X_{i+1}) - \Abar)(\thetahat_{i} - \theta_*) \right] \\
    & \leq \frac{\lambda_A}{t} \sum_{i=1}^t \E \left[ \left\| \thetahat_{i} - \theta_* \right\|^2_2 \right] \to 0,
\end{align*}
since $\thetahat_i \to \theta_*$ a.s.-$\Pbb_{\Wcal | \Dcal}$, by Proposition \ref{thm:plsa_as}. Here,
$\lambda_A = \sup_{x \in \Xcal} \left\|W_1\Atilde(x) - \Abar\right\|_2^2 < \infty$, by Assumption \ref{asu2ii} and the boundedness of $W$.

\color{black}
Now consider the term within the sum in $I_2$. Without loss of generality, assume $i > j$. Let $\Fcal_j$ denote the natural filtration with respect to the Markov chain $\{X_k\}$, upto index $j$. Then, we have
\begin{align*}
    & \quad \E\left[\left\langle \left(W_{i+1} \Atilde(X_{i+1}) - \Abar\right)\left(\thetahat_{i} - \theta_*\right), \left(W_{j+1} \Atilde(X_{j+1}) - \Abar\right)\left(\thetahat_{j} - \theta_*\right) \right\rangle \right] \\
    & \leq \frac{R_\theta^2}{i^\gamma j^\gamma} \E \left[\left\langle W_{i+1} \Atilde(X_{i+1}) - \Abar, W_{j+1} \Atilde(X_{j+1}) - \Abar \right\rangle \right] \\
    & = \frac{R_\theta^2}{i^\gamma j^\gamma} \E \left[ \E\left[ \langle W_{i+1} \Atilde(X_{i+1}) - \Abar, W_{j+1} \Atilde(X_{j+1}) - \Abar\rangle \left| \Fcal_{j+1} \right. \right] \right] \\
    & = \frac{R_\theta^2}{i^\gamma j^\gamma} \E \left[ \left\langle \E\left[ W_{i+1} \Atilde(X_{i+1} \left| \Fcal_{j+1} \right. \right] - \Abar, W_{j+1} \Atilde(X_{j+1}) - \Abar \right\rangle \right],
\end{align*}
where the first step uses $\left\| \thetahat_i - \theta_* \right\| \leq R_\theta i^{-\gamma}$, for some $\gamma \in (0,\eta - 1/2)$ and $R_\theta < \infty$, by Proposition \ref{thm:plsa_as}, while the second step follows from the tower property, conditioning on the filtration $\Fcal_{j+1}$. 

Proceeding from here, we have
\begin{align*}
    & \quad \frac{R_\theta^2}{i^\gamma j^\gamma} \E \left[ \left\langle \E\left[ W_{i+1} \Atilde(X_{i+1}) \left| \Fcal_{j+1} \right. \right] - \Abar, W_{j+1} \Atilde(X_{j+1}) - \Abar \right\rangle \right] \\
    & \leq \frac{R_\theta^2}{i^\gamma j^\gamma} \E \left[ \left\|\E\left[ W_{i+1} \Atilde(X_{i+1}) \left| \Fcal_{j+1} \right. \right] - \Abar\right\|_F \left\|W_{j+1} \Atilde(X_{j+1}) - \Abar \right\|_F \right] \\
    & \leq \frac{R_\theta^2}{i^\gamma j^\gamma} \E \left[ \left\|\E\left[ W_{i+1} A(X_{i+1}) \left| \Fcal_{j+1} \right. \right] - \Abar \right\|_F \left( \left\|W_{j+1} A(X_{j+1}) \right\|_F + \left\|\Abar \right\|_F \right) \right] \\
    & \leq \frac{(1 + W_{\max})A_{\max}R_\theta^2}{i^\gamma j^\gamma} \E \left[ \left\|\E\left[ W_{i+1} A(X_{i+1}) \left| \Fcal_{j+1} \right. \right] - \Abar \right\|_F \right] \\
    & \leq (1 + W_{\max})A_{\max}^2 R_\theta^2 M \frac{\kappa^{i-j}}{i^\gamma j^\gamma}.
\end{align*}
We first bound the inner product using Frobenius norms. In the third step, we bound the second term within the expectation using Assumption \ref{asu2ii} and the boundedness of $W_j$. The final step follows from Lemma \ref{lem:unif_mixing}.

So far, we have shown that
\begin{align*}
    I_2 \leq (1 + W_{\max})A_{\max}^2 R_\theta^2 M \cdot \frac{1}{t} \sum_{i \neq j}^t \frac{\kappa^{i-j}}{i^\gamma j^\gamma}.
\end{align*}
Consider the double sum above. Grouping terms by $l = |i-j|$, we have
\begin{align*}
    \sum_{i \neq j}^t \frac{\kappa^{i-j}}{i^\gamma j^\gamma} = 2\sum_{l=1}^{t-1} S_{t,l} \kappa^l, \quad \text{where} \quad S_{t,l} = \sum_{j=1}^{t-l} \frac{1}{j^\gamma (j+l)^\gamma}.
\end{align*}
Then $S_{t,l} \leq \sum_{j=1}^{t-l} \frac{1}{j^{2 \gamma}}$. For any fixed $l$, $\lim_{t \to \infty} \sum_{j=1}^{t-l}\frac{1}{j^{1 + 2 \gamma}} < \infty$, and so $\lim_{t \to \infty} \frac{1}{t}\sum_{j=1}^{t-l}\frac{1}{j^{2 \gamma}} = 0$, by Kronecker's lemma. Hence, $S_{t,l}/t \to 0$, as $t \to \infty$. Then, by the Dominated Convergence Theorem, we have $\lim_{t \to \infty} \frac{1}{t}\sum_{l=1}^{t-1} S_{t,l} \kappa^l = 0$. It follows that $I_2 \to 0$ as $t \to \infty$, and so $\frac{1}{\sqrt t}\sum_{i=1}^t(W_{i+1} \Atilde(X_{i+1}) - \Abar)(\thetahat_{i} - \theta_*) = o_p(1)$. This concludes the proof. \qed

\vspace{0.3cm}
\color{black}
The following is a restatement of Lemma 2.11 from \cite{vaart1998}.

\begin{lemma}
\label{lem:vaart}
Suppose that $X_n \implies X$ for a random vector $X$ with a continuous distribution function. Then $\sup_x \left|P(X_n \leq x) - P(X \leq x) \right| \to 0$.
\end{lemma}

\subsubsection*{Proof of Theorem \ref{thm:plsa3}}
Let $f(x) = \Atilde(x)\theta_* - \btilde(x)$. Then, by Assumption \ref{asu2ii}, $f$ is bounded, and $$\lim_{t \to \infty}\E[f(X_t)] = \Abar \theta_* - \bbar = 0$$ under the stationary distribution $\mu$.

By the Poisson equation (see e.g., \cite{douc2018}), there exists a bounded function $u$ such that
\begin{align*}
    u(x) - \Pcal u(x) = f(x).
\end{align*}
For $t \geq 0$, we define the following terms:
\begin{align*}
    e_{t+1} & = u(X_{t+1}) - \Pcal u(X_{t}), \\
    r_{t+1} & = \Pcal u(X_{t}) - \Pcal u(X_{t+1}).
\end{align*} 
Let $\Fcal_t = \sigma(\{X_i\}_{i=1}^t)$ denote the natural filtration induced by the Markov chain $\{X_t\}$. Then $f(X_t) = e_t + r_t$, where $e_t$ is a martingale difference sequence, since 
\begin{align*}
    \E[e_{t+1} | \Fcal_{t}] & = \E[u(X_{t+1}) \mid \Fcal_t] - \Pcal u(X_t) = 0,
\end{align*}
and
\begin{align}
    \frac{1}{\sqrt t}\sum_{i=1}^t r_i & = \frac{1}{\sqrt t}(\Pcal u(X_0) - \Pcal u(X_t)) \to 0 \quad a.s., \label{eq:thm_2_3_rk}
\end{align}
as $t \to \infty$, by a telescoping sum argument. Then from (\ref{eq:theorem_2_2b}) we have
\begin{align}
    \sqrt{t}(\bar{\theta}_t - \theta_*) & = -\frac{1}{\sqrt t} \Abar^{-1} \sum_{i=1}^t f(X_{i+1}) + o_p(1) \nonumber \\
    & = -\frac{1}{\sqrt t}\Abar^{-1} \sum_{i=1}^t e_{i+1} + o_p(1), \label{eq:thm_4_2_45}
\end{align}
by (\ref{eq:thm_2_3_rk}). Combined with Proposition \ref{thm:lsa_clt}, this implies that 
\begin{align}
    \frac{1}{\sqrt t}\Abar^{-1} \sum_{i=1}^t e_{i+1} \implies \Ncal(0, \Abar^{-1} Q (\Abar^{-1})\Tra). \label{eq:martingale_clt}
\end{align}
On the other hand, since $e_{i+1}$ is uniformly bounded (as $f(x)$ is uniformly bounded for all $x \in \Xcal$), the Lindenberg condition is satisfied, that is,
\begin{align*}
    \sum_{i=1}^t \E\left[ \frac{\|e_i\|_2^2}{t} I_{\left\{ \|e_i\|_2/\sqrt{t} \geq \epsilon \right\}} \left|\Fcal_{i-1} \right. \right] \to 0,
\end{align*}
in probability, as $t \to \infty$. So, by the martingale central limit theorem (e.g., Lemma A.3. of \cite{liang2010}), we have 
\begin{align}
    \frac{1}{\sqrt t}\Abar^{-1} \sum_{i=1}^t e_{i+1} \implies \Ncal(0, \Lambda), \label{eq:martingale_clt2}
\end{align}
where $\Lambda$ is a positive definite matrix with
\begin{align}
    \Abar^{-1} \sum_{i=1}^t \E[e_i e_i\Tra / t | \Fcal_{i-1}] \left(\Abar^{-1}\right)\Tra \to \Lambda, \label{eq:martingale_clt_Lambda}
\end{align}
in probability as $t \to \infty$. It follows from (\ref{eq:martingale_clt}) and (\ref{eq:martingale_clt2}) that $\Lambda = \Abar^{-1} Q \left( \Abar^{-1} \right)\Tra$, and so, by (\ref{eq:martingale_clt_Lambda}), we have
\begin{align}
    \sum_{i=1}^t \E[e_i e_i\Tra / t | \Fcal_{i-1}] \to Q, \label{eq:martingale_variance}
\end{align}
in probability, as $t \to \infty$.

Next, from (\ref{eq:theorem_2_2c}), we have 
\begin{align*}
     \sqrt{t} (\bar{\thetahat}_t - \bar{\theta}_t) & = -\frac{1}{\sqrt t} \Abar^{-1} \sum_{i=1}^t (W_{i+1} - 1)f(X_{i+1}) + o_p(1) \\
     & = -\frac{1}{\sqrt t} \Abar^{-1}\sum_{i=1}^t (W_{i+1} - 1)e_{i+1} + o_p(1),
\end{align*}
using (\ref{eq:thm_2_3_rk}). Let $\xi_t = (W_{t} - 1)e_{t}$. Then $\xi_t$ is a martingale difference sequence, since
\begin{align*}
    \E[\xi_{t+1} \mid \Fcal_t] = \E[W_{t+1}-1]\E[e_{t+1} \mid \Fcal_t] = 0.
\end{align*}
Since $\xi_t$ is uniformly bounded, the Lindenberg condition holds. Then, by the martingale central limit theorem, conditional on the data $\Dcal$, the term $\frac{1}{\sqrt t}\Abar^{-1} \sum_{i=1}^t \xi_{i+1}$ is asymptotically normal with mean 0 and variance
\begin{align*}
    \plim_{t \to \infty} \Abar^{-1} \sum_{i=1}^t \E[\xi_i \xi_i\Tra / t | \Fcal_{i-1}] \left( \Abar^{-1} \right)\Tra & = \plim_{t \to \infty} \Abar^{-1} \Var(W_1) \sum_{i=1}^t \E[e_i e_i\Tra / t | \Fcal_{i-1}] \left( \Abar^{-1} \right)\Tra \\
    & = \Abar^{-1} Q \left( \Abar^{-1} \right)^{-1},
\end{align*}
where the first equality follows by independence of $W_i$ and $e_i$ and the fact that the $W_i$'s are i.i.d., while the second equality follows from (\ref{eq:martingale_variance}) and $\Var(W_1) = 1$. So, we have
\begin{align}
\label{eq:plsa_clt}
    \sqrt{t}(\bar{\thetahat}_t - \bar{\theta}) = -\frac{1}{\sqrt t}\Abar^{-1}\sum_{i=1}^t \xi_{i+1} + o_p(1) \implies \Ncal(0,\Abar^{-1} Q (\Abar^{-1})\Tra),
\end{align}
as $t \to \infty$, where the asymptotic normality holds conditional on data $\Dcal$. 

Let $X \sim \Ncal(0, \Abar^{-1} Q (\Abar^{-1})\Tra)$ denote the random variable with the limiting distribution of $\sqrt{t}(\thetabar_t - \theta_*)$ as $t \to \infty$. Applying Lemma \ref{lem:vaart} to the result of Proposition \ref{thm:lsa_clt} and equation (\ref{eq:plsa_clt}), respectively, we get
\begin{align*}
    \sup_{v \in \R^d}\left| \Pbb_\Dcal(\sqrt{t}(\thetabar_t - \theta_*) \leq v) - \Pbb(X \leq v) \right| & \to 0, \text{and} \\
    \sup_{v \in \R^d}\left|\Pbb_{\Wcal | \Dcal}(\sqrt{t}(\bar{\thetahat}_t - \theta_*) \leq v) - \Pbb(X \leq v)\right| & \to 0, \textrm{in probability},
\end{align*}
as $t \to \infty$. Then
\begin{align*}
   & \sup_{v \in \R^d}\left|\Pbb_{\Wcal | \Dcal}(\sqrt{t}(\bar{\thetahat}_t - \theta_*) \leq v) - \Pbb_\Dcal(\sqrt{t}(\thetabar_t - \theta_*) \leq v)\right| \\& \leq \sup_{v \in \R^d}\left| \Pbb_{\Dcal}(\sqrt{t}(\thetabar_t - \theta_*) \leq v) - \Pbb(X \leq v)\right| \\ & + \sup_{v \in \R^d}\left|\Pbb_{\Wcal | \Dcal}(\sqrt{t}(\bar{\thetahat}_t - \theta_*) \leq v) - \Pbb(X \leq v)\right| \\ 
    & \to 0,
\end{align*}
in probability, as $t \to \infty$. \qed

\color{black}
\section{Additional Experiments}
\label{sec:appendix_experiment}
In this section, we provide the study of the second-order accuracy of our bootstrap method in the Frozenlake environment considered in Section \ref{sec:frozenlake}.

\begin{figure}[h!]
    \centering
    \begin{subfigure}{0.485\textwidth}
        \includegraphics[width=\hsize]{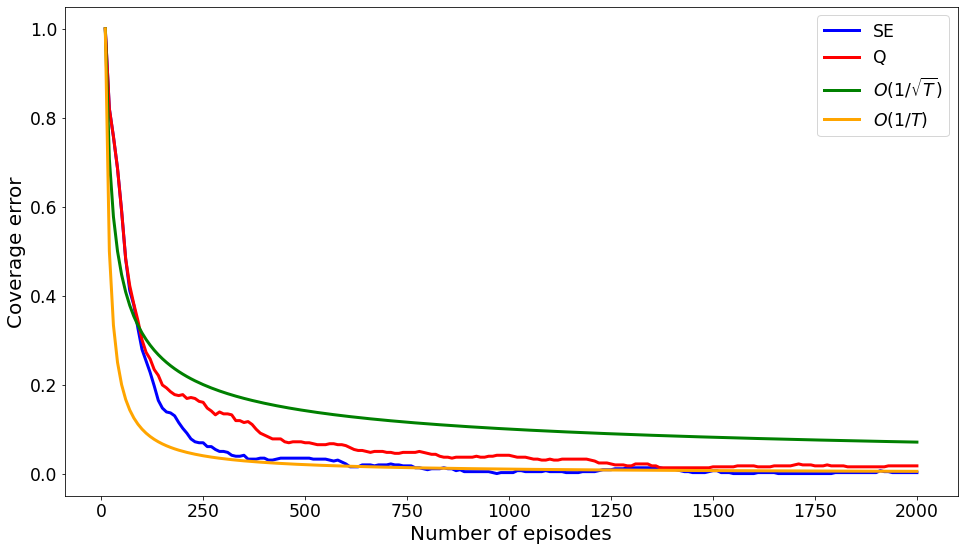}
        \caption{Coverage error rates}
        \label{fig:frozenlake_coverage_error}
    \end{subfigure}%
    \hspace{0.02\textwidth}%
    \begin{subfigure}{0.485\textwidth}
        \includegraphics[width=\hsize]{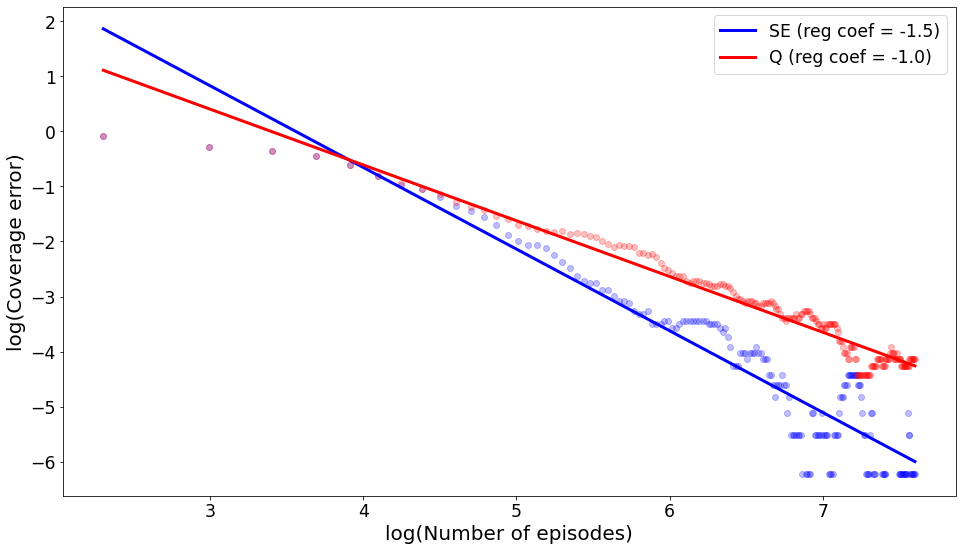}
        \caption{Least squares estimates}
        \label{fig:frozenlake_coverage_error_reg}
    \end{subfigure}%
    \caption{Figure \ref{fig:frozenlake_coverage_error} shows the coverage error rates for the quantile and SE confidence intervals, and Figure \ref{fig:frozenlake_coverage_error_reg} shows the linear regression coefficients for the log coverage error rates against the log number of episodes.}
    \label{fig:frozenlake_coverage_error_plots}
\end{figure}

To empirically evaluate the second-order accuracy, we measured the coverage error rates of the 95\% confidence intervals for the value function of the initial state in the Frozenlake environment. We use TD learning to estimate the value function, and the quantile and standard error estimators, computed from the online bootstrap estimates, to generate the confidence intervals. Figure \ref{fig:frozenlake_coverage_error} shows the empirical coverage errors of the quantile and standard error estimators as a function of the number of episodes in the RL Frozenlake environment. The rates are re-scaled to start from 1 at the first time step. In both cases, we can see that the coverage error decreases at a rate faster than $O(1/\sqrt{t})$ initially, and eventually reaches a rate of $O(1/t)$ or better.

We then computed estimates of the coverage error rate by regression the log of the coverage errors against the log of the number of episodes. We would expect a first-order accurate method to have a regression coefficient of $-1/2$ or lower (corresponding to a coverage error rate of $O(1/\sqrt{t})$), while a second-order accurate method would have a coefficient of $-1$ or lower. As shown in Figure \ref{fig:frozenlake_coverage_error_reg}, both the quantile and standard error have regression coefficients of -1 or lower, which demonstrates that they both achieved second-order accuracy.

\end{document}